\documentclass[review]{elsarticle}

\usepackage{lineno,hyperref}
\modulolinenumbers[5]

\journal{Journal of \LaTeX\ Templates}

\usepackage{indentfirst}
\usepackage{amsmath}
\usepackage{lineno}
\usepackage[justification=centering]{caption}
\usepackage{caption}
\usepackage{slashbox,multirow}
\usepackage{algpseudocode}
\usepackage{CJK}
\usepackage{indentfirst}
\usepackage{rotating}
\usepackage{graphicx}
\usepackage{booktabs}
\usepackage{array}
\usepackage{multirow}
\usepackage{booktabs}
\usepackage{pifont}
\usepackage{changepage}
\usepackage{amssymb}
\usepackage{bigdelim,multirow}
\usepackage{latexsym}
\usepackage{epstopdf}
\usepackage{booktabs}
\usepackage{threeparttable}
\usepackage{setspace}
\usepackage{mdframed}
\usepackage{diagbox}
\usepackage[ruled,vlined]{algorithm2e}
\usepackage{subfigure}
\usepackage{ragged2e}
\usepackage{caption}
\usepackage{float}
\usepackage[graphicx]{realboxes}
\usepackage{color}
\usepackage{url}
\usepackage{mathtools}
\usepackage{verbatim}
\usepackage{ragged2e}
\usepackage{hyperref}
\usepackage[section]{placeins}
\renewcommand{\raggedright}{\leftskip=0pt \rightskip=0pt plus 0cm}
\usepackage{ragged2e}

\begin{document}

\begin{frontmatter}

\title{A Self-adaptive Weighted Differential Evolution Approach for Large-scale Feature Selection}

\author[1]{Xubin Wang}
\ead{wangxb19@mails.jlu.edu.cn}
\author[2]{Yunhe Wang\corref{mycorrespondingauthor}}
\ead{wangyh082@hebut.edu.cn}
\author[3]{Ka-Chun Wong}
\ead{kc.w@cityu.edu.hk}
\author[1]{Xiangtao Li\corref{mycorrespondingauthor}}
\cortext[mycorrespondingauthor]{Corresponding author}
\ead{lixt314@jlu.edu.cn}
\address[1]{School of Artificial Intelligence, Jilin University, Changchun, China}
\address[2]{School of Artificial Intelligence, Hebei University of Technology, Tianjin, China. }
\address[3]{Department of Computer Science, City University of Hong Kong, Hong Kong}

\begin{abstract}
Recently, many evolutionary computation methods have been developed to solve the feature selection problem. However, the studies focused mainly on small-scale issues, resulting in stagnation issues in local optima and numerical instability when dealing with large-scale feature selection dilemmas. To address these challenges, this paper proposes a novel weighted differential evolution algorithm based on self-adaptive mechanism, named SaWDE, to solve large-scale feature selection. First, a multi-population mechanism is adopted to enhance the diversity of the population. Then, we propose a new self-adaptive mechanism that selects several strategies from a strategy pool to capture the diverse characteristics of the datasets from the historical information. Finally, a weighted model is designed to identify the important features, which enables our model to generate the most suitable feature-selection solution. We demonstrate the effectiveness of our algorithm on twelve large-scale datasets. The performance of SaWDE is superior compared to six non-EC algorithms and six other EC algorithms, on both training and test datasets and on subset size, indicating that our algorithm is a favorable tool to solve the large-scale feature selection problem. Moreover, we have experimented SaWDE with six EC algorithms on twelve higher-dimensional data, which demonstrates that SaWDE is more robust and efficient compared to those state-of-the-art methods. SaWDE source code is available on Github at https://github.com/wangxb96/SaWDE.
\end{abstract}


\end{frontmatter}


\section{Introduction}
%
%
%
%
In the context of the rise of Internet technology, data is growing exponentially, leading to the accumulation of numerous large-scale datasets \cite{cai2018feature}. Hence, it is essential to obtain valuable information from such huge amount of data \cite{ayesha2020overview}. However, the increasing amount of high-dimensional data brings the “curse of dimensionality” problems, which poses computational challenges especially for classification \cite{xue2019self} \cite{song2020variable}. In addition, existing computational algorithms reveal some limitations, such as high complexity, computationally expensive, poor robustness, and low generalizability \cite{wainwright2019high}. Therefore, it is necessary to develop effective computational models to select important features in data classification to discover valuable information \cite{A17}.

Feature selection \cite{A18} is an effective method to reduce redundant features and has been proved useful in the data process, regarded as a combinatorial optimization problem, with $2^n-1$ solutions for $n$-dimensional features \cite{xue2019self}. In the past, many heuristic methods have been proposed to address feature selection problems which can be divided into three categories \cite{sheikhpour2017survey}, including filter methods, wrapper methods and embedded methods. Filter methods \cite{labani2018novel} can uncover the importance of features using the internal structure of the training data, which takes less time in the training stage; for instance, Liu \textit{et al.} \cite{liu1996probabilistic} proposed a probabilistic-based filter solution for feature selection. Yu \textit{et al.} \cite{yu2003feature} developed a fast correlation-based filter solution to tackle the high-dimensional feature selection problem. Hancer \textit{et al.} \cite{hancer2018differential} combined the information theory and feature ranking with DE for feature selection. However, filter methods cannot always provide the good results. The wrapper method \cite{sheikhpour2016particle} adopts the performance of the last-used learner as the evaluation criterion for feature subsets, while the embedded method \cite{zhang2015embedded} automatically selects features in the learning and training process through a combination of filter method and wrapper method; for instance, Kabir \textit{et al.} \cite{kabir2010new} proposed a neural network-based wrapper feature selection method. Mafarja \textit{et al.} \cite{mafarja2018whale} applied the whale optimization algorithm to feature selection. Maldonado \textit{et al.} \cite{maldonado2009wrapper} proposed a SVM-based wrapper approach. Wang \textit{et al.} \cite{wang2015embedded} proposed an embedded unsupervised feature selection method. Maldonado \textit{et al.} \cite{maldonado2018dealing} developed an embedded strategy that penalizes the cardinality of the feature set by a scaling factor technique. Lu \textit{et al.} \cite{lu2019embedded} proposed a new embedded methods considering unknown data heterogeneity. However, these heuristic methods can easily trapped in the local optima due to the uncertainty of selecting and evaluating features individually by greedy strategies. Moreover, the performance of these heuristics is not sufficiently stable since their performance often depends on the particular scenario and the design approach.

To improve the quality of feature selection and to avoid stagnation in local optimality, evolutionary computation (EC) methods \cite{A3}, including  Genetic algorithms (GAs) \cite{sayed2019nested} , particle swarm optimization (PSO) \cite{zhang2017pso}, ant colony optimization (ACO) \cite{manoj2019aco} , artificial bee colony algorithm (ABC) \cite{hancer2018pareto},  and differential evolutionary (DE) \cite{zhang2020binary} algorithms have been proposed to address the aforementioned problems. For example, Chen \textit{et al.} \cite{chen2020evolutionary} proposed a multitasking-based evolutionary feature selection approach for high-dimensional classification. Sayed \textit{et al.} \cite{sayed2019nested} developed a Nested-GA method to find the optimal feature subset in high-dimensional cancer Microarray datasets. Xue \textit{et al.} \cite{xue2012particle} designed a multi-objective based PSO feature selection approach. Tran \textit{et al.} \cite{tran2017new} proposed a potential particle swarm optimization (PPSO) algorithm with a new representation method for feature selection. Xue \textit{et al.} proposed a novel initialization and mechanism in PSO for feature selection. Kashef \textit{et al.} \cite{kashef2015advanced} presented an advanced binary ACO (ABACO), which treats each feature as a graph node and regards the feature selection problem as a graph model. Hancer \textit{et al.} integrated a similarity search strategy in the ABC algorithm \cite{hancer2015binary}. Mlakar \textit{et al.} \cite{mlakar2017multi} proposed a multi-objective differential evolution feature selection method for facial expression recognition. Zorarpac \textit{et al.} \cite{zorarpaci2016hybrid} combined ABC and DE algorithm to construct a hybrid method for feature selection. Zhang \textit{et al.} \cite{zhang2020binary} proposed a binary DE algorithm with self-learning strategy for multi-objective feature selection problem. Khushaba \textit{et al.} \cite{khushaba2011feature} proposed a repair mechanism to integrate with DE for feature selection.

 
Although many EC algorithms have been employed to address feature selection problems, most EC methods always encounter the complication of stagnation in local optima and numerical instability when dealing with large-scale feature selection problems since large-scale data contains more irrelevant and redundant features \cite{xue2012particle} \cite{chen2020evolutionary}. The reason may be that many EC methods are unable to explore and exploit the search space in a balanced manner under different conditions. Recently, the self-adaptive mechanism has been proved as an effective strategy for solving feature selection. For instance, Xue \textit{{et al.}} \cite{xue2020self} proposed a self-adaptive strategy-based PSO algorithm to exploit the global and local information for feature selection. Aladeemy \textit{{et al.} }\cite{aladeemy2017new} proposed the self-adaptive cohort intelligence (SACI) for coinstantaneous feature selection and model selection. Xue \textit{et al.} \cite{xue2019self} developed a self-adaptive PSO feature selection method for large-scale datasets. Xue \textit{et al.} \cite{xue2014ensemble} proposed a self-adaptive learning techniques-based ensemble algorithm for high-dimensional numerical optimization. Brester \textit{ {et al.} }\cite{brester2014self} investigated several multi-objective genetic algorithms for choosing the most important features in a dataset. Huang \textit{{et al.}} \cite{huang2014music} used a self-adaptive harmony search (SAHS) algorithm to select local feature subsets to achieve an automatic music genre-classification system. Essentially, we observe that the algorithms are able to address the problems that standard EC methods cannot.

Differential evolution algorithm is a heuristic stochastic search algorithm for solving optimization problems based on population differences, which was proposed by R. Storn and K. Price with the advantages of fast convergence, few control parameters, and simple setup \cite{storn1997differential}. The differential evolution algorithm evolves multiple solutions by mutation, crossover, and selection, searching for the best solution. Several studies have already applied DE to feature selection; for instance, Zainudin \textit{{et al.} }\cite{zainudin2017feature} combined relief-f with DE for feature selection, in which a self-adaptive mechanism can adjust the population and generation size. Aladeemy \textit{{et al.}} \cite{aladeemy2020new} proposed an opposition-based self-adaptive cohort intelligence (OSACI) algorithm \cite{aladeemy2017new}. Ghosh \textit{{et al.}} \cite{ghosh2013self} proposed self-adaptive differential evolution (SADE) to generate feature subsets. Besides, Fister \textit{{et al.}} \cite{fister2018novel} used a threshold mechanism in self-adaptive differential evolution (SADE) to eliminate the irrelevant features. Gaspar-Cunha \textit{{et al.}} \cite{gaspar2014self} designed a self-adaptive evolutionary multi-objective approach (MOEA), in which the parameters of the classifier are dynamically updated. However, most of these studies lacked generalization ability and neglected to consider high-dimensional data.

Meanwhile, since large-scale data has more search space, an improper search may lead to a time consuming and low classification performance. Hence, population partitioning techniques have been developed to diversify candidate solutions and strategies to find the best answer for large-scale feature selection problems. Zhang \textit{{et al.}} \cite{zhang2019novel} proposed a multi-population niche GA (MPNGA) for feature selection. It combines several filter methods and basic knowledge to reduce the barriers for enhancing search ability of multi-populations. The experiment conducted in this paper shows that the structure of the multi-populations is useful for keeping the population diversified. Park \textit{{et al.}} \cite{park2020multi} proved that a multi-population approach can prevent premature convergence in the course of evolution. Besides, Chen \textit{{et al.}} \cite{chen2020efficient} designed a multi-population original fruit fly algorithm (MOFOA) to boost the search ability and improve performance of feature selection results. Meanwhile, Nseef \textit{{et al.}} \cite{nseef2016adaptive} proposed an adaptive multi-population ABC algorithm for dynamic optimization problems to maintain diversity and cope with dynamic changes. In addition, Chen \textit{{et al.}} \cite{chen2020multi} combined DE with three different embedded multi-population mechanisms for Harris hawk hunting optimization, showing that the method can effectively enhance exploratory and exploitative performance.

From this perspective, we designed a weighted self-adaptive DE algorithm for feature selection. In this algorithm, a self-adaptive search mechanism from global to local is proposed, and five equal sub-populations are generated. Besides, a pool of candidate solution generation strategies is constructed to find the most suitable evolutionary strategy in a dynamic way. Then, eight mutation strategies are considered and the five best-performing mutation strategies are selected to further form the strategy pool. Further, a weighted model is designed to identify the most important features, enabling the model to generate the best solution. The proposed SaWDE algorithm is tested on twelve benchmark datasets and compared to several benchmark algorithms. Results indicate an effective competitive performance compared to these algorithms. Moreover, we have experimented SaWDE with six EC algorithms on twelve higher-dimensional data, which demonstrates that SaWDE is more robust and efficient compared to those state-of-the-art methods. 

The rest of the paper is organized as follows: Section \uppercase\expandafter{\romannumeral2}  describes the basic knowledge of original DE. Section \uppercase\expandafter{\romannumeral3} describes the proposed method in detail. Section \uppercase\expandafter{\romannumeral4} details the experimental design. Section \uppercase\expandafter{\romannumeral5} shows the experimental results with discussion. In Section \uppercase\expandafter{\romannumeral6} , we draw the final conclusions and present future work.

\section{Differential Evolution} 
Differential evolution algorithm, proposed by R. Storn and K. Price \cite{storn1997differential}, is a stochastic heuristic search algorithm for solving optimization problems. In the beginning, each candidate solution can be denoted as $P_{i} = \{ p_i^1,p_i^2,...,p_i^D\}$, where $i$ = 1, 2, ..., $N$, and $D$ is the dimension of the data. $P_i$ is initialized randomly as follows:
\begin{equation}
    \Vec{p_{i,G}^j} = \Vec{p_{min}^j} + rand(0,1) \cdot (\Vec{p_{max}^j} - \Vec{p_{min}^j})~~~j = 1, 2, ..., D
\end{equation}
where $P_{min} = \{p_{min}^1, p_{min}^2, ...,p_{min}^D\}$ and $P_{max} = \{p_{max}^1, p_{max}^2, ...,p_{max}^D\}$, $G$ is the number of generations and $rand(0,1)$ denotes a random number on the interval [0,1]. 

After that, DE algorithm achieves individual mutation by differential mutation strategy. Taking the mutation strategy ``DE/best/1" as an example, the newly generated mutation vector is as follows:
\begin{equation}
    \Vec{V_{i,G+1}} = \Vec{P_{best,G}} + F \cdot (\Vec{P_{r_1,G}} - \Vec{P_{r_2,G}})
\end{equation}
where $r_1$ and $r_2$ are two mutually exclusive random numbers within the range [1, $N$], $\vec{P_{best,G}}$ is the best individual in the generation $G$, and $F$ is the scaling factor  for scaling the difference vector.

Then, the DE algorithm randomly selects individuals by the crossover operation, and the trail vector $\Vec{U_{i,G}}=\{u_{i,G}^1,u_{i,G}^2,...,u_{i,G}^D\}$ is generated as follows:
\begin{equation}
    \Vec{u_{i,G}^j} = \begin{dcases}
    \Vec{v_{i,G}^j}, ~~~ if~ rand(0,1) \leq CR \\
    \Vec{p_{i,G}^j} ~~~~ otherwise
    \end{dcases}
\end{equation}
where CR is the crossover probability. After that, the selection operation is applied to select the better individual as follows: 
\begin{equation}
    \Vec{P_{i,G}} = \begin{dcases}
    \Vec{U_{i,G}}, ~~~ if~ f(U_{i,G}) \leq f(X_{i,G})  \\
    \Vec{P_{i,G}} ~~~~ otherwise
    \end{dcases}
\end{equation}
where $f$ denotes the objective function.

\section{Methods}
\subsection{Methodology Overview of SaWDE}
In this study, we develop the SaWDE algorithm to find the best feature subsets on large-scale data. The schematic overview of the algorithm is summarized in Fig. 1. And the main thoughts of SaWDE can be seen from Algorithm 1. First of all, the multi-population mechanism is employed to increase the diversity of a population. In this experiment, the original population is divided into five same sized sub-populations, each of which chooses specific solution generation strategies through the self-adaptive mechanism. The evolution of each sub-population is carried out separately, and a search operation is carried out on each sub-population. Moreover, to maintain the diversity of each sub-population, its individuals are dynamically changed at each generation. 

\begin{figure*}[htb]
\centering
\includegraphics[scale = 0.36]{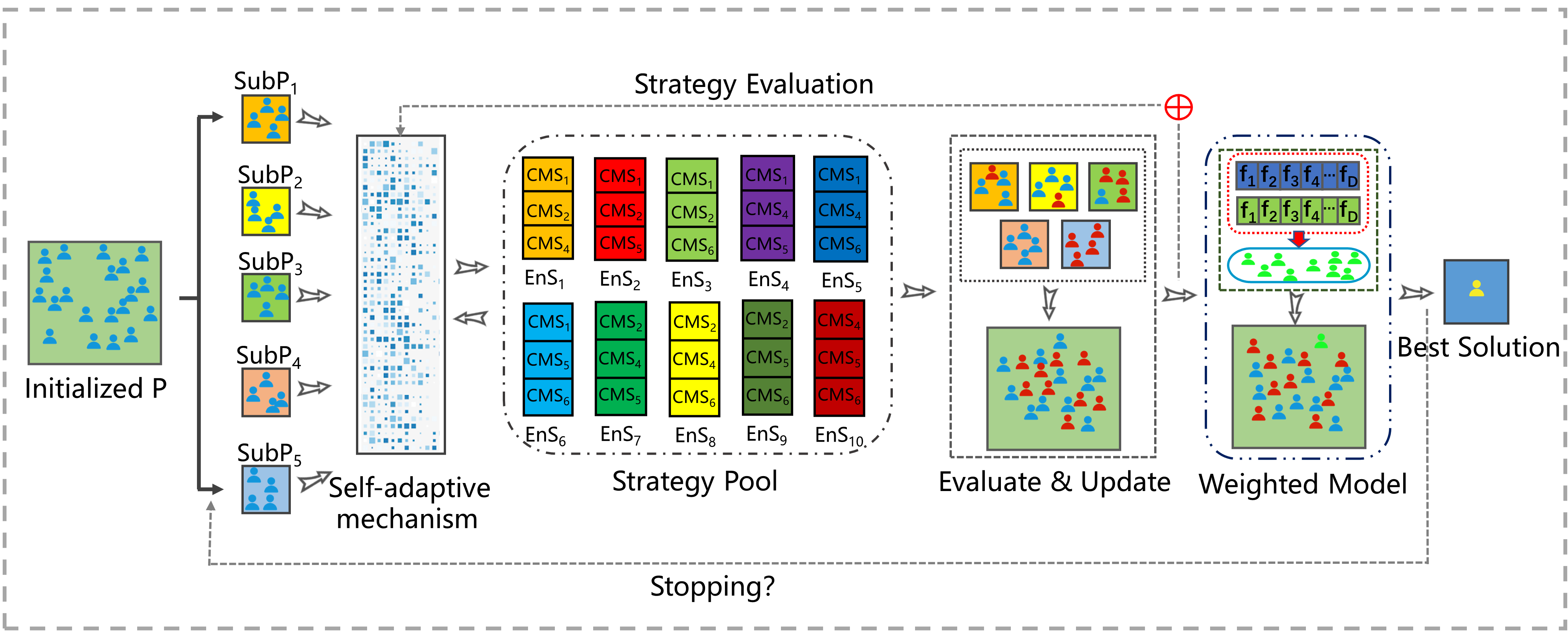}
\caption{\raggedright Overview of the proposed SaWDE algorithm. The initialized population $P$ is divided equally into five sub-populations, $SubP_1$, $SubP_2$, $SubP_3$, $SubP_4$, $SubP_5$. Then, each sub-population selects an strategy \emph{EnS} through the self-adaptive mechanism from the strategy pool. After that, the $P$ is evaluated and updated by the selected \emph{EnS}. Meanwhile, the strategies \emph{EnSs} are evaluated for additional rewards and further sub-strategy pool construction. Finally, a weighted model is proposed to assess the importance of each feature and search for the best solution by evaluating these features in a combinatorial way. }
\label{fig:graph}
\end{figure*}

Second, a strategy pool including ten types of mechanism is employed in evolution. We named those mechanism as $EnS_1$ to $EnS_{10}$, where each strategy contains three different single candidate mutation scenarios (\emph{CMS}). In particular,  \emph{EnS}$_1$ includes \emph{CMS}$_1$, \emph{CMS}$_2$ and \emph{CMS}$_4$, \emph{EnS}$_2$ includes \emph{CMS}$_1$, \emph{CMS}$_2$ and \emph{CMS}$_5$, \emph{EnS}$_3$ includes \emph{CMS}$_1$, \emph{CMS}$_2$ and \emph{CMS}$_6$, \emph{EnS}$_4$ includes \emph{CMS}$_1$, \emph{CMS}$_4$ and \emph{CMS}$_5$, \emph{EnS}$_5$ includes \emph{CMS}$_1$, \emph{CMS}$_4$ and \emph{CMS}$_6$, \emph{EnS}$_6$ includes \emph{CMS}$_1$, \emph{CMS}$_5$ and \emph{CMS}$_6$, \emph{EnS}$_7$ includes \emph{CMS}$_2$, \emph{CMS}$_4$ and \emph{CMS}$_5$, \emph{EnS}$_8$ includes \emph{CMS}$_2$, \emph{CMS}$_4$ and \emph{CMS}$_6$, \emph{EnS}$_9$ includes \emph{CMS}$_2$, \emph{CMS}$_5$ and \emph{CMS}$_6$, and \emph{EnS}$_{10}$ includes $\emph{CMS}_4$, \emph{CMS}$_5$ and \emph{CMS}$_6$. \emph{CMSs} represent DE mutation strategies, which are detailed in Section 3.4. After that, strategies are first selected in a random way and the performance of each strategy is recorded. Then, to select strategies from the the strategy pool in a self-adaptive manner, a self-adaptive selection mechanism is proposed. After that, the strategy pool is further reduced according to the performance of each previous strategy and used to bring the selection from global to local, which ensures the diversity of strategies and enhances the searching for ability of the best-performing strategy. Moreover, during this evolution course, the self-adaptive mechanism can conduct additional incentive selection each twenty generations to further improve the search ability of our algorithm, inspired by \cite{wu2016differential}.
\\

\begin{algorithm}[H]
\scriptsize
\captionsetup{font={scriptsize}}
\caption{Pseudo Code of the SaWDE Algorithm.}
\KwIn{Set population ($P$), number of population (N), the dimension of data (D), fitness evaluations (FES), maximum FES (MaxFES), subset size (SZ), candidate mutation scenario (CMS), classification accuracy (Acc), the objective function ($f$) and so on.
 }
\KwOut{ The classification accuracy $f(P_i)$ and the subset size;
}

Diversification initialization\;
$f(P) \leftarrow f(P_i, P_{i+1}, ..., P_{N})$\;
\While{(FES $< =$ MaxFES)}{
$[SubP_1, SubP_2, SubP_3, SubP_4, SubP_5]\leftarrow$ Randomly partition $P$ into 5 sub-populations\;
    \For {i = 1 $\rightarrow$  5} {
        $SubP_i \leftarrow$ Use the \emph{ Self-adaptive strategy mechanism} to select $Strategy_j$\;
        \For {m = 1 $\rightarrow$ 3} {
        $SubP_i$ $\leftarrow$ $CMS(m)$ in $Strategy_j$\;
            \For{ k = 1 $\rightarrow$ N/5} {
            $f(SubP_i)$ $\leftarrow$ $f(SubP_{i(1)}, SubP_{i(2)}..., SubP_{i(N/5)})$, Update individuals;
            }
        }
        Update FES\;
        weight $\leftarrow$Use the \emph{weighted model} to calculate the feature importance\;
        Use the \emph{self-adaptive strategy mechanism} to evaluate the performance of the selected $Strategy$\;
    }
    $P\leftarrow$ Update the $P$ by \emph{weighted model}\;
}
\end{algorithm}

Finally, a weighted model is proposed to discover the important features of the dataset at each generation. In this model, we first record the updated features of the current population and those features in the top 20\% of individuals in the sub-populations. After that, the model selects and evaluates the features in a combinatorial way according to their rank determined in the previous step.

\subsection{Representation of Solutions}
Since the standard DE algorithm is a continuous optimization algorithm, the standard continuous encoding scheme of DE cannot be used to directly address large-scale feature selection problems. In our study, we transfer a continuous vector into a binary string under a threshold $\theta$. At first, to apply an individual to represent the feature, the set population, $P$ with $N$-dimensional vector can be defined as follows:

\begin{equation}
    P = \{P_1, P_2, ..., P_N\}
\end{equation}

\begin{equation}
    P_i = \{p_i^1, p_i^2, ..., p_i^D\}, \quad   i = 1, 2, ..., N
\end{equation}

After that, we use the threshold $\theta$, to transfer each element of the individual $P_i$ into a binary string. If the value of the $j$th dimension of the individual $P_i$ is greater than $\theta$, we set $p_i^j$ to 1. Otherwise, the $j$th dimension of $P_i$ is set as 0. We can observe that the value in $P^j$ is 0 or 1. 1 represents that the $j$th feature is selected, while 0, means that the $j$th feature is not selected, which can be described as follows:

\begin{equation}
 P_i^j =\begin{dcases}
1,&p_i^j \geq \theta\\
0,&p_i^j < \theta
\end{dcases}
\end{equation}

\subsection{Multi-population-based Strategy}

In this work, the population $P$ will be randomly and dynamically generated with equal-sized sub-populations. To maintain the diversity of populations without increasing the complexity of our algorithm, five random sub-populations $SubP_1$, $SubP_2$, $SubP_3$, $SubP_4$ and $SubP_5$ will be generated in each iteration, as follows: 
\begin{equation}
    P = \{SubP_1, SubP_2, SubP_3, SubP_4, SubP_5\}
\end{equation}

We use $N$, as the population size of the parent population, and $N_1$, $N_2$, $N_3$, $N_4$, and $N_5$, represent the size of the sub-populations $SubP_1$, $SubP_2$, $SubP_3$, $SubP_4$ and $SubP_5$ respectively. In our study, each sub-population has an equal population size.
\begin{equation}
    N = \sum_{i = 1}^{5}N_i
\end{equation}

\begin{equation}
    N_1 = N_2 = N_3 = N_4 = N_5
\end{equation}

Next, each sub-population is independently optimized for each generation in turn. During the evolution, a sub-population selects its own strategy separately by self-adaptive mechanism, and then evaluates and evolves individuals in each subspace. As each sub-population adopts the self-adaptive mechanism to select particular strategies, different strategies may be used in different sub-populations, which increases the exploration ability of the algorithm from several perspectives. 

In addition, the size of sub-population will not change at each generation. Moreover, the individuals can evolve dynamically in each generation, maintaining the diversity of the sub-population and avoiding getting trapped in local optima.

\subsection{Construction of the Mutation Strategy Pool}
In DE, particular mutation strategies have different performances on the various datasets and as such, some mutation strategies may be more appropriate at specific stages of the evolution than the single mutation strategy. So, we select multiple strategies to accelerate the convergence speed by establishing a strategy pool. Before constructing the strategy pool, selecting appropriate mutation strategies is a very important step due to their potential different performance.

There are two aspects to consider, how many mutation strategies should be selected to build the pool, and which should be selected. Based on this, we first chose eight typical mutation operators that are representative of the current DE algorithms, to build eight candidate mutation strategies \emph{CMSs}. In our investigation, \emph{CMS}$_1$ to \emph{CMS}$_8$ represent, 
``DE/current to best/1", ``DE/current to rand/1", ``DE/rand/3", ``DE/best/1", ``DE/rand to best/1", ``DE/rand/2", ``DE/best/2" and ``DE/best/3" \cite{wang2011differential}\cite{qin2008differential}.

During the evolution, DE generates a mutation vector $\Vec{V_{i,G}}$ = ($v_{i, 1, G}, v_{i, 2, G}, \\..., v_{i, D ,G}$), for each individual $\Vec{P_{i,G}}$ in the $G$-th generation. The indices $r1, r2, r3,\\ r4, r5, r6$ and $r7$ are random integers, which are mutually exclusive within the range [1, N]. The $\Vec{P_{best,G}}$, denotes the best individual in the present population. Besides, $F$ is the scaling factor for scaling the difference vector.
    
\begin{itemize}
\item [1)]``DE/current to best/1"
      \begin{equation}
          \Vec{V_{i,G}} = \Vec{P_{i,G}} + F \cdot (\Vec{P_{best,G}} - \Vec{P_{i,G}}) + F \cdot (\Vec{P_{r1,G}} - \Vec{P_{r2,G}}).
      \end{equation}
\item [2)]``DE/current to rand/1"
      \begin{equation}
          \Vec{V_{i,G}} = \Vec{P_{i,G}} + rand \cdot (\Vec{P_{r1,G}} - \Vec{P_{i,G}}) + F \cdot (\Vec{P_{r2,G}} - \Vec{P_{r3,G}}).
      \end{equation}
\item [3)]``DE/rand/3"
      \begin{gather}
          \Vec{V_{i,G}} = \Vec{P_{r1,G}} +  F \cdot (\Vec{P_{r2,G}} - \Vec{P_{r3,G}} + \Vec{P_{r4,G}} - \Vec{P_{r5,G}} +  \Vec{P_{r6,G}} - \Vec{P_{r7,G}}).    
      \end{gather}

\item [4)]``DE/best/1"
      \begin{equation}
          \Vec{V_{i,G}} = \Vec{P_{best,G}} + F \cdot (\Vec{P_{r1,G}} - \Vec{P_{r2,G}}).
      \end{equation}
\item [5)]``DE/rand to best/1"
      \begin{equation}
          \Vec{V_{i,G}} = \Vec{P_{r1,G}} + F \cdot (\Vec{P_{best,G}} - \Vec{P_{i,G}}) + F \cdot (\Vec{P_{r2,G}} - \Vec{P_{r3,G}}).
      \end{equation}
\item [6)]``DE/rand/2"
      \begin{equation}
          \Vec{V_{i,G}} = \Vec{P_{r1,G}} + F \cdot (\Vec{P_{r2,G}} - \Vec{P_{r3,G}}) + F \cdot (\Vec{P_{r4,G}} - \Vec{P_{r5,G}}).
      \end{equation}
\item [7)]``DE/best/2"
      \begin{equation}
          \Vec{V_{i,G}} = \Vec{P_{best,G}} + F \cdot (\Vec{P_{r1,G}} - \Vec{P_{r2,G}}) + F \cdot (\Vec{P_{r3,G}} - \Vec{P_{r4,G}}).
      \end{equation}
\item [8)]``DE/best/3"
      \begin{gather}
          \Vec{V_{i,G}} = \Vec{P_{best,G}} + F \cdot (\Vec{P_{r2,G}} - \Vec{P_{r3,G}} + \Vec{P_{r4,G}} - \Vec{P_{r5,G}} + \Vec{P_{r6,G}} - \Vec{P_{r7,G}}).          
      \end{gather}
\end{itemize}

Following, as practice is the only criterion for testing truth, we identify effective \emph{CMSs} experimentally. First, the strategies are tested respectively on a variety of datasets. Then, the top five \emph{CMSs} are selected to form the strategy pool. 

To enhance the search ability and prevent overfitting, we combine every three different \emph{CMSs} to form the ensemble mechanism as a final strategy, called \emph{EnS}, five \emph{CMSs} are selected and ten different \emph{EnSs} generated according to the combination principle. Specifically, \emph{EnS}$_1$ includes \emph{CMS}$_1$, \emph{CMS}$_2$ and \emph{CMS}$_4$, \emph{EnS}$_2$ includes \emph{CMS}$_1$, \emph{CMS}$_2$ and \emph{CMS}$_5$, \emph{EnS}$_3$ includes \emph{CMS}$_1$, \emph{CMS}$_2$ and \emph{CMS}$_6$, \emph{EnS}$_4$ includes \emph{CMS}$_1$, \emph{CMS}$_4$ and \emph{CMS}$_5$, \emph{EnS}$_5$ includes \emph{CMS}$_1$, \emph{CMS}$_4$ and \emph{CMS}$_6$, \emph{EnS}$_6$ includes \emph{CMS}$_1$, \emph{CMS}$_5$ and \emph{CMS}$_6$, \emph{EnS}$_7$ includes \emph{CMS}$_2$, \emph{CMS}$_4$ and \emph{CMS}$_5$, \emph{EnS}$_8$ includes \emph{CMS}$_2$, \emph{CMS}$_4$ and \emph{CMS}$_6$, \emph{EnS}$_9$ includes \emph{CMS}$_2$, \emph{CMS}$_5$ and \emph{CMS}$_6$, and \emph{EnS}$_{10}$ includes $\emph{CMS}_4$, \emph{CMS}$_5$ and \emph{CMS}$_6$. These \emph{EnSs} compose our initial strategy pool, which can seen in Fig. 2.

\begin{figure*}[!htb]
\centering
\includegraphics[scale = 0.36]{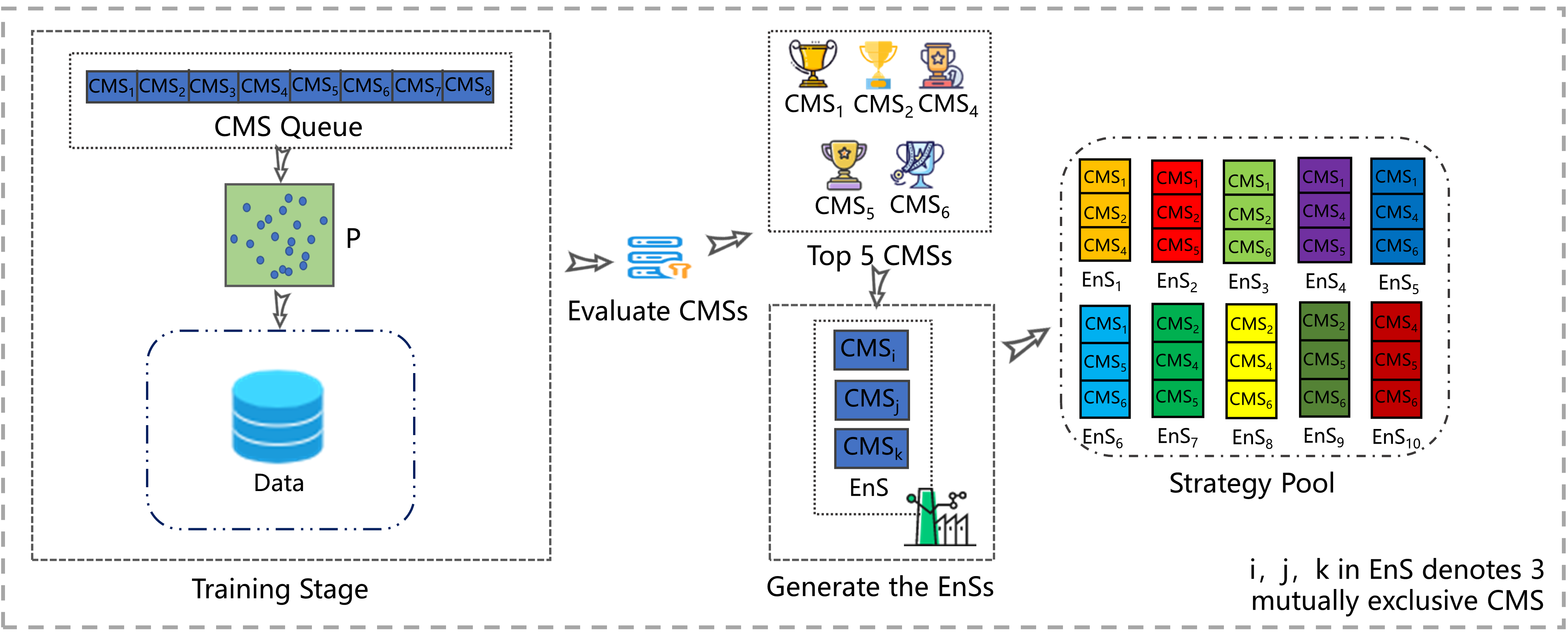}
\caption{\raggedright The process of strategy pool construction. First, the \emph{CMSs} are in a queue and selected to train the training datasets in sequence. After that, all the \emph{CMSs} are evaluated based on their performance, and the top 5 \emph{CMSs} are selected to construct the strategy pool. On this basis, every three different \emph{CMSs} are selected to construct an \emph{EnS} ensemble strategy and all of these make up the initial strategy pool. }
\label{fig:graph}
\end{figure*}

However, once a strategy pool is built it usually does not change. Some strategies \emph{EnSs} in the strategy pool will not play an active role in the evolution process, which increases the computational cost of the algorithm and does not improve the algorithm's performance. Moreover, a single strategy is insufficient to meet the requirements of evolution at different stages due to lack of diversity as stated above. Therefore, we propose a self-adaptive mechanism to choose and evaluate the \emph{EnSs}.

\subsection{Self-Adaptive Mechanism}
In our study, the adaptive parameters in our algorithm are F and CR, similar to the references \cite{wu2016differential} \cite{zhang2009jade}. Under the adaptive F and CR, we propose our self-adaptive mechanism to automatically adapt the most
appropriate \emph{EnS} depends on the characteristics of the datasets and their performance. There are two aspects to consider in self-adaptive mechanism: (1) How to choose the \emph{EnSs} ensemble strategy; (2) and how to evaluate each \emph{EnS} during evolution. Moreover, the performance of the DE is strongly related to the setting of the strategy and control parameters \cite{mallipeddi2011differential}. To address these problems, a novel self-adaptive mechanism is proposed here so that choosing the most appropriate \emph{EnS} depends on the characteristics of the datasets and and the \emph{EnSs}' performance on these datasets. The proposed self-adaptive mechanism is described in Algorithm 2.
 
First of all, the self-adaptive selection mechanism randomly chooses \emph{EnSs} during the first half iterations to maintain a fair competition. Thereafter, we use the $EnSNum$ to record the number of each \emph{EnSs} selected. Meanwhile, we also record the increased accuracy as, $Change$ for each \emph{EnSs} that improved the individuals in the population successfully, which may be defined as follows:
 
\begin{equation}
    EnSNum(EnS_i) = EnSNum(EnS_i) + 1,\quad  i = 1, 2, ..., 10
\end{equation}
 
\begin{gather}
     Change(EnS_i) = Change(EnS_i) + \tau, \quad i = 1, 2, ..., 10     
\end{gather}
where, '$\tau$' denotes
the increased accuracy of successful individuals. After each twenty generations, the \emph{EnSs} with the best performance is rewarded regarding its prior performance, $\psi$. The $\psi$ is defined by the maximum ratio of the increased accuracy, $Change$ with the consumed FES ($cFES$). Then, the $Reward$ for each \emph{EnSs} is: 
\begin{equation}
    \psi(EnS_i) = Change(EnS_i) / cFES(EnS_i),\quad i = 1, 2, ..., 10
\end{equation}

\begin{algorithm}[H]
\scriptsize
\captionsetup{font={scriptsize}}
\caption{Pseudo Code of the Self-adaptive Strategy Mechanism.}
\KwIn{sub population ($SubP_i$), consumed FES ($cFES$), the increased accuracy of changed individuals ($Change$), count the number of selected \emph{EnSs} ($EnSNum$), the number of reward \emph{EnSs} ($Reward$), $SR$ is used to analyze the relationship between strategy choice and reward strategy, $SPro$ is the top five strategies according to $SR$, and $\psi$ denotes the change rate in accuracy of individuals.
}
\KwOut{The selected \emph{EnS}.
}
    \While{(FES $< =$ MaxFES)}{
       \For{i = 1 $\rightarrow{}$ 5 }{
         \If{ FES $<=$ $\frac{1}{2}$ MaxFES}{
            $SubP_i$ $\leftarrow{}$ $Random (EnS_j)$;}
         \Else{
           $SR = Reward ./ EnSNum$\; 
           $SR$ = $Sort(SR, 'descend')$\;
           $SPro$ $\xleftarrow{}$ $SR_{1  \xrightarrow{} 5}$\;
           $SubP_i$ $\xleftarrow{}$ $Random (SPro)$\;}
           \If{$mod (generations, 20) == 0$}{
              $\psi = Change / cFES$\;
              $EnS_j$ = $Max (\psi)$\;
              $SubP_i$ $\leftarrow{}$ \emph{EnS}$_j$\;
              $Reward(EnS_j)$=$Reward(EnS_j)$+1\;}
           $f(SubP_i)$ $\leftarrow{}$ $f(P_1, P_2, ... ,P_k)$\;
           Update $EnSNum, Change$ and FES\;
        }
       Update FES\;
   }
\end{algorithm}

\begin{figure*}[!htp]
\centering
\includegraphics[scale = 0.36]{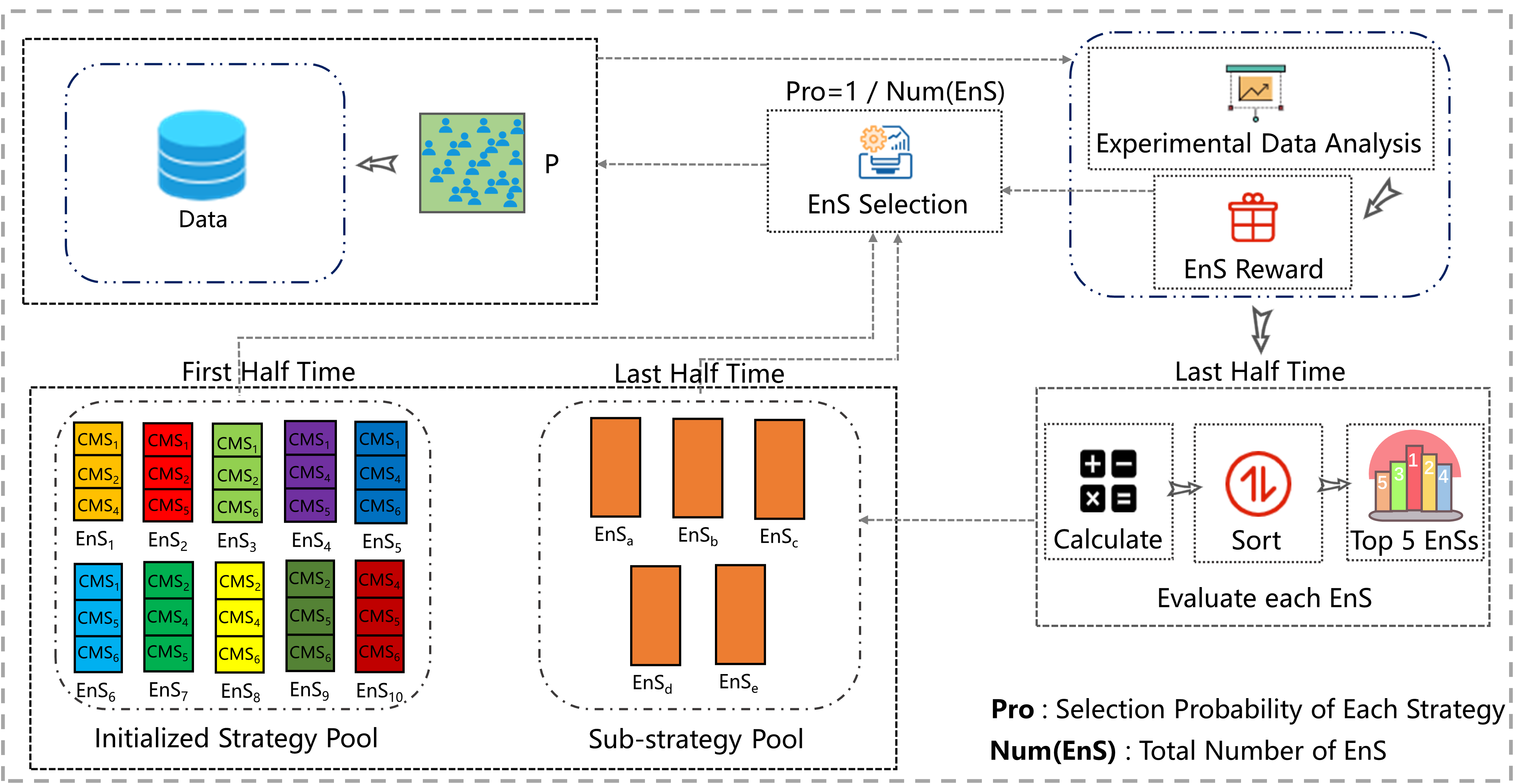}
\caption{\raggedright The workflow of the self-adaptive mechanism. Throughout evolution, each \emph{EnS} in the strategy pool is selected with the same probability in the first half time. The selected \emph{EnS} is then evaluated and recorded. After that, the best \emph{EnS} on each twenty generation performance will give the extra selection reward. The strategy pool is the same as the initialized strategy pool in Fig. 2 in the first half time but reduced by half according to the performance of \emph{EnSs} in the last half time. The top 5 \emph{EnSs} are dynamically selected to the sub-strategy pool before each function evaluation based on their performance. Then those \emph{EnSs} are selected in the same way as the first half time.} 
\label{fig:graph}
\end{figure*}

\begin{gather}
    Reward(EnS_i) = Reward(EnS_i) + 1, 
    \quad i = 1, 2, ..., 10    
\end{gather}

Based on this, the \emph{EnSs} are selected from global to local based on the performance of first half time. Specifically, a sub-strategy pool with half the content of the original strategy pool is constructed to update the strategy search from global to local. To build that, we first calculate the performance of each strategy \emph{EnS} before each evaluation, then the top five \emph{EnSs} with the highest ratio of total number of rewards (\emph{Reward}) to total number of choices (\emph{EnSNum}) are selected to construct the sub-strategy pool for further search. After that, the \emph{EnS} is selected from the sub-strategy pool in the same way as the first half time to ensure the appropriate \emph{EnSs} are selected.

\subsection{Weighted Model}
To assess the importance of the each feature, a weighted model is proposed to calculate the weight of features in evolution. There are two main processes in the weighted model, the first one records and assesses the importance of each feature at each generation, and the second searches for the solution feature subset.

\begin{figure}[!htb]
\centering
\includegraphics[scale = 0.52]{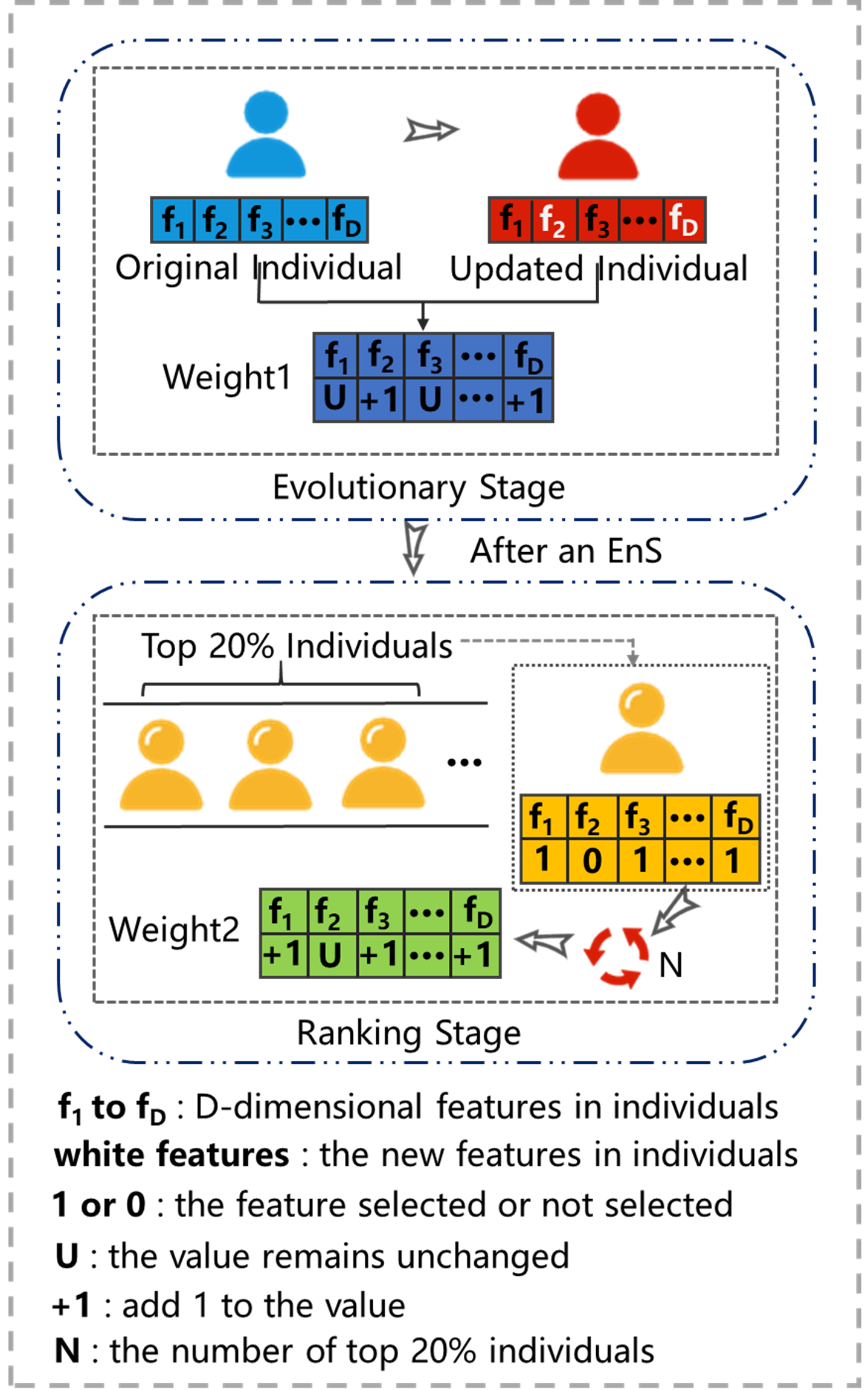}
\caption{\raggedright The recording procedure of the weighted model. During the evolutionary stage, the new features in individual are recorded by a matrix $Weight1$, when it updates successfully. And after an \emph{EnS}, all the individuals are ranked due to their performance, and the top 20\% individuals' all features are recorded by a matrix $Weight2$. }
\label{fig:graph}
\end{figure}

The first process consists of two steps each having a different assessment approach of important features. The first step (CF) works in the evolutionary stage, it stores the new features of individuals who performed better during evolution, and the second step records directly all the features of the best 20\% of individuals (AF) at the end of an \emph{EnS} in the ranking stage. The first 2-step process can be formulated as follows:

\begin{equation}
    weight1(f_i) = weight1(f_i) + CF, \quad i = 1, 2, ..., D
\end{equation}
\begin{equation}
    weight2(f_i) = weight2(f_i) + AF,\quad i = 1, 2, ..., D
\end{equation}

The aim is to rank the importance of the features and then reduce the search space to find a good solution efficiently and effectively. A more vivid view of the first weighting process is depicted in Fig. 4.

In the second process, solutions are searched every twenty generations considering the first process. The features are first selected according to their rank in the corresponding weight evaluation matrix, according to the two different weightings of process 1 and then each feature subset sequentially evaluated at each stage. After that, the best individual is compared with the worst individual in the parent population, and if the selected best individual shows a better performance, replaces that individual. The procedure of the weighted model can be seen in Algorithm 3.

\subsection{Time Complexity Analysis}
In this section, we analyze the time complexity of SaWDE method. In the initialization phase, SaWDE costs $O(N \times D)$, where $N$, is the number of individuals in the population and $D$, is the dimension of the dataset. After that, as each dataset is classified via KNN, it costs $O(N \times (n \times D + K \times n))$, where $n$, is the number of samples and $K$, is the parameter used to determine the number of nearest neighbors. On evolution, SaWDE costs $O(N \times D)$ in the transformation from real numbers to 0 and 1 before function evaluation, and costs $O(N \times (n \times D + K \times n))$ for function evaluation in each generation. In addition, the self-mechanism costs, $O(MaxFES)$. What's more, the weighted model costs $O(N \times D)$ in the first step and $O(t \times j)$ in the second step of the first process at each generation, where $t$, is the number of changed individuals and $j$, is the number of updated features. In the second process, the weighted model costs, $O(D^2)$ for searching the potential solutions and costs, $O(D \times (n \times D + K \times n))$ for function evaluations.  

\begin{algorithm}[H]
\scriptsize
\captionsetup{font={scriptsize}}
  \caption{Pseudo Code of the Weighted Model.}
  \KwIn{selected features in individual ($SF$), updated features in individual ($UF$), sub population ($SubP_i$), $weight1$ $\xleftarrow{}$ zeros (1, D), $weight2$ $\xleftarrow{}$ zeros (1, D).   }
  \KwOut{A new $P$.}
    
    \While{ FES $<$ MaxFES}{
         \For{i = 1 $\xrightarrow{}$ 5 }{
             $SubP_i$  $\xleftarrow{}$ $EnS$\;
             $f (SubP_i)$ $\xleftarrow{}$ $f (P_{l   \xrightarrow{} N(SubP_i)}) $\;
             \For{each k in $UF$}{
             $weight1 (1, UF (k))$$\xleftarrow{}$$weight1(1, UF(k))$+1\; }
             $f (SubP_i)$ $\xleftarrow{}$ $sort (f (SubP_i),  'descend')$\;
             \For{k = 1 $\xrightarrow{}$ $\frac{1}{5}$ * $f$ ($SubP_i$ ($f$ ($P_{l \xrightarrow{} N(SubP_i)}$)))}{
                \For{each j in $SF$}{
                  $weight2(1,SF (j))\xleftarrow{}$$weight2 (1,SF(j))$ + 1\; }
             }
            } 
        \If{$mod$ (generations, 20) == 0}{
           $SelectFeature_1$ = $sort(weight1, 'descend')$\;
           $SelectFeature_2$ = $sort(weight2, 'descend')$\;
           \For{ i = 1 $\xrightarrow{}$ $\frac{1}{2}$ D}{
                $Y_i$ $\xleftarrow{}$ $Y_{i} (SelectFeature_1 (1 \xrightarrow{} i)$\; 
                $f (Y)$ $\xleftarrow{}$ $f (Y_i)$\;}
            \For{ i = $\frac{1}{2}$  D + 1 $\xrightarrow{}$ D}{
               $Y_i$ $\xleftarrow{}$ $Y_{i} (SelectFeature_2 (1 \xrightarrow{} i - \frac{1}{2}D)$\; 
               $f (Y)$ $\xleftarrow{}$ $f (Y_i)$\;}
            \If{$Max(f(Y)) > Min(f(P)) $}{
               $Min(f(P)) = Max(f(Y))$ ;}
        }
         Update FES\;
   }
\end{algorithm}

\section{Experimental Design}

Our experimental strategy is first to construct a good strategy pool for SaWDE by selecting the five best performers from eight candidate \emph{CMSs}. Meanwhile, every three \emph{CMSs} from the five selected \emph{CMSs} are combined into an \emph{EnS} to make up a robust strategy for evolution. After that, the performance of SaWDE is compared to the above eight \emph{CMSs}. 
  
Second experiment we test which population size ($N$) is best for the main experiment. The influence of different population size on evaluation performance is different, a smaller population is not conducive to global search and a larger population will increase the cost of calculation. Therefore, a suitable parent population size is a useful preparation for evolution. In this experiment, we set the N to 50, 100, 200, 300 or 500, on all twelve datasets. 

The main experiment is to test the performance of the SaWDE algorithm on the twelve datasets. After the experiment, it will compare other compared algorithms with the performance on training and test datasets.

Finally, without the loss of generality, twelve higher-dimensional datasets are used for evaluating the SaWDE model.

\subsection{Datasets}

As datasets, we use twelve datasets from the University of California Irvine (UCI) Machine Learning Repository \cite{xue2019self}. Each dataset contains information about instances, features and classes. The details of the datasets are shown in Table 1 $Dataset ~{\uppercase\expandafter{\romannumeral1}}$. Each dataset was randomly divided into a training dataset and test dataset with a ratio of 7 to 3, respectively. 

To further validate the performance of SaWDE, twelve higher-dimensional datasets \cite{de2008clustering} are employed to evaluate SaWDE, which can download from ``\url{https://schlieplab.org/Static/ Supplements/ CompCancer/datasets.htm}''. The details of these data are summarized in Table 1, which contains a wide range of cancer types, mostly above 1000 dimensions, with the highest dimensionality of 4553.

\begin{table}[!htb]
\centering
\caption{High-dimensional datasets used to evaluate the performance of SaWDE. Each dataset contains information about instances, features and classes.}
\resizebox{123mm}{!}{
\begin{tabular}{|l l l l||l l l l|}
\hline
\textbf{Dataset \uppercase\expandafter{\romannumeral1}} & \textbf{Instances} & \textbf{Features} & \textbf{Classes}&\textbf{Dataset \uppercase\expandafter{\romannumeral2}} & \textbf{Instances} & \textbf{Features} & \textbf{Classes}\\
\hline
grammatical$\_$facial$\_$expression01 & 1,062 & 301 & 2 & Alizadeh-2000-v1 & 42 & 1095 & 2\\
SemeionHandwrittenDigit & 675 & 256 & 10 & Alizadeh-2000-v2 & 62 & 2093 & 3\\
 isolet5 & 1,040 & 617 & 26  & Armstrong-2002-v1 &	72	&	1081	&	2  \\
 MultipleFeaturesDigit & 1000 & 649 & 10 & Bittner-2000 & 38	& 2201	& 2\\
 HAPTDataSet & 1200 & 561 & 12 & Dyrskjot-2003 & 40	& 1203 & 3\\
har & 900 & 561 & 6 & Garber-2001 &	66	&	4553	&	4\\
 UJIIndoorLoc & 900 & 522 & 3 & Liang-2005 &	37	&	1411	&	3\\
 MadelonValid & 600 & 500 & 2 & Nutt-2003-v2 &	28	& 1070 & 2\\
OpticalRecognitionofHandwritten & 1000 & 64 & 10 & Pomeroy-2002-v1 &	34	&	857	&	2\\
 ConnectionistBenchData & 208 & 60 & 2 & Pomeroy-2002-v2 & 42 &	1379 &	5\\
 wdbc & 596 & 30 & 2 & Shipp-2002-v1 &	77	&	798	&	2 \\
 LungCancer & 32 & 56 & 3 & West-2001 &	49	& 1198	&	2\\
\hline

\end{tabular}}
\end{table}

\subsection{Parameter Settings}
We use MaxFES (Maximum Function Evaluations) as the stop criterion. In all experiments, we set the MaxFES to $1 \times 10^6$. Meanwhile, SaWDE ends the evolution early when 100\% accuracy is achieved in training and the feature subset size is less than half of the original subset. The parameter settings of the comparison algorithms are summarized in Table 2. For our proposed SaWDE, we set population size $N$ to 100, and eight initial F and CR are set to 0.5, 1, 0.6, 0.9, 0.5, 0.9, 0.6 and 1; 0.1, 0.2, 0.9, 0.8, 0.9, 0.1, 0.8 and 0.2, respectively. KNN is used to classify all datasets over the evolution process, the number of nearest neighbors $K$ is set to 3 \cite{xue2015survey}. Meanwhile, we used the 3-fold cross-validation in KNN \cite{xue2019self}. 

\begin{table}[!htb]
\centering
\caption{Parameter values of the comparison algorithms \cite{xue2019self}}
\resizebox{88mm}{!}{
\begin{tabular}{l l}
\hline
\textbf{Algorithms} & \textbf{Parameter values}  \\
\hline 
LRS21 & l = 2; r = 1  \\[3pt]
LRS32 & l = 3; r = 2  \\[3pt]
DE & F = 0.5; CR = 0.1; MaxFES = $1 \times 10^6$ \\[3pt]
SaDE & Initial CR = 0.5; $F \in N(0.5, 0.3^2)$; LP = 10; MaxFES = $1 \times 10^6$ \\[3pt]
GA & CR = 0.7; MR = 0.1;  SR = 0.5; MaxFES = $1 \times 10^6$ \\[3pt]
Original PSO & C1 = C2 = 1.49618;  w = 0.7298; MaxFES = $1 \times 10^6$  \\[3pt]
Standard PSO & C1 = C2 = 1.49618;  $w \in [0.9,0.4]$; MaxFES = $1 \times 10^6$  \\[3pt]
SaPSO & $p_j$ = 0.2; LP = 10; ps = 100; $\theta$ = 0.6 ; Ub = 1; Lb = 0;\\[3pt] & Ubv = 0.5; Lbv = -0.5; MaxFES = $1 \times 10^6$  \\[3pt]
SaWDE & Initial CR = [0.1, 0.2, 0.9, 0.8, 0.9, 0.1, 0.8, 0.2]; \\[3pt]& Initial F = [0.5, 1, 0.6, 0.9, 0.5, 0.9, 0.6, 1]; MaxFES = $1 \times 10^6$ \\[3pt]
\hline
\end{tabular}}
\end{table}

\section{Results}

\subsection{Computational Results and Comparisons}
In this part, we compare the experimental results of our SaWDE algorithm with those six other non-EC and six EC algorithms. The results are summarized as the four parts, namely, the subset size, classification accuracy and convergence curves of the algorithm on the training datasets, the classification accuracy on the test datasets. As can be seen from the experimental results, the performance of our SaWDE algorithm is better than that of all the EC and non-EC algorithms.

\subsubsection{Results of subset size on Test Datasets}
Table 3 shows the subset sizes obtained by the SaWDE algorithm and six non-EC and six EC algorithms. The upper half of Table 3 shows the experimental results obtained with the non-EC and SaWDE algorithms, and the lower half Table 3 the experimental results obtained with the EC and SaWDE algorithms. In the table, the first column represents the data used in the experiment, and the subset sizes are given as mean values of the trials after all experimental times,  `\%' represents the reduction rate between the subset size and the original size. The best results for each dataset are highlighted in bold text.

In the non-EC comparison, we see that although our SaWDE algorithm doesn't necessarily perform the best in percentage reduction, it has a reduction rate of more than 90\% or close to this in over half of the datasets. At the same time, the SaWDE algorithm works nearly as well as the best results for subset sizes on many datasets.

In the EC algorithm comparison, SaWDE's subset size results from all but the eleventh dataset are much better than other EC algorithms. Further, the advantages of SaWDE in subset size become more apparent especially as the feature dimensions of the dataset get larger and larger. For example, compared to SaPSO, the best previous EC algorithm, SaWDE reduces the subset size by 10 to 20 percent more on most datasets. In addition, as seen in the Table, the standard deviation of our algorithm on solving the subset size is relatively low, indicating the stability of our algorithm.

\begin{table*}[htb]
\centering
\caption{Subset sizes of six other non-EC, six EC methods and SaWDE on training datasets}
\resizebox{125mm}{!}{
\begin{tabular}{l l l l l l l l l l l l l l l l  }
\hline
 \multirow{2}{*}{Datasets} & \textbf{SFS} & \textbf{ } & \textbf{SBS} & \textbf{ } & \textbf{LRS21} & \textbf{ } & \textbf{LRS32} & \textbf{ } & \textbf{SFFS} & \textbf{ } & \textbf{SBFS} & \textbf{ } & \textbf{SaWDE} & \textbf{ }\\
 \cline{2-16}
   \textbf{ } & \textbf{Mean $\pm$ Std} & \textbf{\%} & \textbf{Mean $\pm$ Std} & \textbf{\%} & \textbf{Mean $\pm$ Std} & \textbf{\%} & \textbf{Mean $\pm$ Std} & \textbf{\%} & \textbf{Mean $\pm$ Std} & \textbf{\%} & \textbf{Mean $\pm$ Std} & \textbf{\%} & \textbf{Mean $\pm$ Std} & \textbf{\%} \\
\hline 
 grammatical$\_$facial$\_$expression01 & 4.9 $\pm$ 1.4 & 98.3 & 298.3 $\pm$ 1.4 & 0.8 & \textbf{4.8} $\pm$ 1.1 & 98.4 & 5.6 $\pm$ 1.3 & 98.1 & 5.7 $\pm$ 1.6 & 98.1 & 298.4 $\pm$ 0.9 & 0.8 & 7.5 $\pm$ 1.0 &97.5  \\[3pt]
 SemeionHandwrittenDigit & 14.8 $\pm$ 5.0 & 94.2 & 254.4 $\pm$ 0.9 & 0.6 & 14.0 $\pm$ 4.0 & 94.5 & 14.1 $\pm$ 3.8 & 94.4 & \textbf{11.8 }$\pm$ 2.4 & 95.3 & 253.9 $\pm$ 0.3 & 0.8 & 79.5 $\pm$ 13.0 & 68.9  \\[3pt]
 isolet5 & 16.0 $\pm$ 3.9 & 97.4 & 615.2 $\pm$ 0.9 & 0.2 & 15.9 $\pm$ 3.0 & 97.4 & 16.3 $\pm$ 3.5 & 97.3 & \textbf{13.6} $\pm$ 2.0 & 97.7 & 614.8 $\pm$ 0.3 & 0.3 & 75 $\pm$ 2.0 & 87.8\\[3pt]
 MultipleFeaturesDigit & 10.2 $\pm$ 2.5 & 98.4 & 646.1 $\pm$ 0.9 & 0.4 & 10.1 $\pm$ 2.7 & 98.4 & \textbf{9.5} $\pm$ 2.4 & 98.5 & 10.8 $\pm$ 1.6 & 98.3 & 646.5 $\pm$ 0.6 & 0.3 & 108.3 $\pm$ 55.5 & 83.3 \\[3pt]
 HAPTDataSet & 8.5 $\pm$ 2.0 & 98.4 & 559.2 $\pm$ 1.1 & 0.3 & \textbf{8.3} $\pm$ 1.6 & 98.5 & 8.7 $\pm$ 2.1 & 98.4 & 9.6 $\pm$ 2.3 & 98.2 & 558.6 $\pm$ 0.6 & 0.4 & 51.5 $\pm$ 23.0 & 90.8 &\\[3pt]
 har & 7.7 $\pm$ 2.3 & 98.6 & 559.3 $\pm$ 1.0 & 0.3 & \textbf{7.1} $\pm$ 1.6 & 98.7 & 7.1 $\pm$ 2.1 & 98.7 & 9.0 $\pm$ 2.3 & 98.3 & 558.9 $\pm$ 0.3 & 0.3 & 60.8 $\pm$ 2.5 & 89.2  \\[3pt]
 UJIIndoorLoc & 1.8 $\pm$ 0.4 & 99.6 & 521.0 $\pm$ 0.0 & 0.1 & 2.0 $\pm$ 0.0 & 99.6 & 2.0 $\pm$ 0.0 & 99.6 & 2.0 $\pm$ 0.4 & 99.6 & 519.8 $\pm$ 0.6 & 0.4 & \textbf{1.0} $\pm$ 0.0 & 99.8 \\[3pt]
 MadelonValid & \textbf{3.1 }$\pm$ 2.4 & 99.3 & 498.1 $\pm$ 0.9 & 0.3 & 5.7 $\pm$ 2.7 & 98.8 & 3.7 $\pm$ 2.5 & 99.2 & 6.5 $\pm$ 2.0 & 98.6 & 497.8 $\pm$ 0.4 & 0.4 &17.8 $\pm$ 7.2 & 96.5 \\[3pt]
OpticalRecognitionofHandwritten & 15.3 $\pm$ 2.5 & 76 & 62.4 $\pm$ 0.8 & 2.4 & 15.4 $\pm$ 3.1 & 75.8 & 14.9 $\pm$ 2.6 & 76.6 & \textbf{9.7} $\pm$ 2.5 & 84.7 & 61.7 $\pm$ 0.6 & 3.5 & 25.8 $\pm$ 5.5 & 59.8 \\[3pt]
 ConnectionistBenchData & 4.1 $\pm$ 1.6 & 93.1 & 58.1 $\pm$ 0.9 & 3 & 4.5 $\pm$ 1.5 & 92.3 & \textbf{3.9} $\pm$ 1.2 & 93.3 & 4.4 $\pm$ 1.6 & 92.6 & 57.7 $\pm$ 0.5 & 3.7 & 8.8 $\pm$ 4.9 & 85.4  \\[3pt]
 wdbc & 3.4 $\pm$ 1.0 & 88.5 & 28.4 $\pm$ 0.7 & 5.2 & 3.5 $\pm$ 1.0 & 88.2 & 3.3 $\pm$ 1.1 & 89 &\textbf{ 3.1} $\pm$ 1.0 & 89.5 & 27.8 $\pm$ 0.3 & 7.1 & 11.3 $\pm$ 4.2 & 62.5 \\[3pt]
 LungCancer & 3.1 $\pm$ 1.4 & 94.4 & 54.1 $\pm$ 0.8 & 3.2 & 2.9 $\pm$ 1.0 &\textbf{ 94.7} & 3.3 $\pm$ 1.4 & 94.1 & 3.5 $\pm$ 1.7 & 93.7 & 53.8 $\pm$ 0.4 & 3.8 & 5.8 $\pm$ 3.2 & 89.7 \\[3pt]
\hline
\end{tabular}}

\centering
\resizebox{125mm}{!}{
\begin{tabular}{ l l l l l l l l l l l l l l l l}
\hline
 \multirow{2}{*}{Datasets} & \textbf{GA} & \textbf{ } & \textbf{Original PSO} & \textbf{ } & \textbf{Standard PSO} & \textbf{ } & \textbf{SaPSO} & \textbf{ } & \textbf{DE} & \textbf{ } & \textbf{SaDE} & \textbf{ } & \textbf{SaWDE} & \textbf{ }\\
 \cline{2-16}
   \textbf{ } & \textbf{Mean $\pm$ Std} & \textbf{\%} & \textbf{Mean $\pm$ Std} & \textbf{\%} & \textbf{Mean $\pm$ Std} & \textbf{\%} & \textbf{Mean $\pm$ Std} & \textbf{\%} & \textbf{Mean $\pm$ Std} & \textbf{\%} & \textbf{Mean $\pm$ Std} & \textbf{\%} & \textbf{Mean $\pm$ Std} & \textbf{\%}\\
\hline 
 grammatical$\_$facial$\_$expression01 & 124.6 $\pm$ 13.8 & 58.6 & 123.7 $\pm$ 10.1 & 58.8 & 121.0 $\pm$ 8.0 & 59.8 & 77.1 $\pm$ 10.1 & 74.3 & 121.7 $\pm$ 8.4 & 59.5 & 114.0 $\pm$ 7.7 & 62.1  & \textbf{7.5} $\pm$ 1.0 &97.5  \\[3pt]
 SemeionHandwrittenDigit & 188.1 $\pm$ 22.5 & 26.5 & 150.0 $\pm$ 13.9 & 41.3 & 165.3 $\pm$ 15.8 & 35.4 & 107.5 $\pm$ 4.8 & 58 & 110.9 $\pm$ 17.2 & 56.6 & 108.3 $\pm$ 7.2 & 57.6 & \textbf{79.5} $\pm$ 13.0 & 68.9  \\[3pt]
 isolet5 & 339.3 $\pm$ 51.6 & 45 & 262.3 $\pm$ 20.8 & 57.4 & 286.9 $\pm$ 36.8 & 53.4 & 159.3 $\pm$ 8.1 & 74.1 & 244.5 $\pm$ 11.4 & 60.3 & 233.6 $\pm$ 9.9 & 62.1 & \textbf{75} $\pm$ 2.0 & 87.8\\[3pt]
 MultipleFeaturesDigit & 333.8 $\pm$ 48.5 & 48.5 & 294.3 $\pm$ 24.6 & 54.6 & 299.9 $\pm$ 24.0 & 53.7 & 147.4 $\pm$ 14.9 & 77.2 & 252.7 $\pm$ 12.7 & 61 & 249.3 $\pm$ 11.2 & 61.5 & \textbf{108.3} $\pm$ 55.5 & 83.3  \\[3pt]
 HAPTDataSet & 324.9 $\pm$ 54.8 & 42 & 273.5 $\pm$ 24.4 & 51.2 & 286.8 $\pm$ 31.2 & 48.8 & 122.9 $\pm$ 15.6 & 78 & 227.0 $\pm$ 11.6 & 59.5 & 220.2 $\pm$ 11.7 & 60.7 &\textbf{ 51.5} $\pm$ 23.0 & 90.8  \\[3pt]
 har & 342.1 $\pm$ 49.9 & 39 & 289.3 $\pm$ 29.2 & 48.4 & 308.9 $\pm$ 34.7 & 44.9 & 123.0 $\pm$ 15.8 & 78 & 224.1 $\pm$ 11.6 & 60 & 216.5 $\pm$ 10.8 & 61.3 & \textbf{60.8} $\pm$ 2.5 & 89.2 \\[3pt]
 UJIIndoorLoc & 225.4 $\pm$ 45.0 & 56.8 & 85.4 $\pm$ 8.8 & 83.6 & 23.0 $\pm$ 6.7 & 95.5 & 3.4 $\pm$ 4.1 & 99.3 & 171.2 $\pm$ 2.9 & 67.1 & 155.7 $\pm$ 4.5 & 70.1 & \textbf{1.0} $\pm$ 0.0 & 99.8 \\[3pt]
 MadelonValid & 290.6 $\pm$ 54.1 & 41.8 & 228.9 $\pm$ 20.6 & 54.2 & 255.9 $\pm$ 38.4 & 48.8 & 111.8 $\pm$ 10.8 & 77.6 & 201.0 $\pm$ 11.6 & 59.7 & 185.8 $\pm$ 10.5 & 62.8 &\textbf{17.8} $\pm$ 7.2 & 96.5 \\[3pt]
 OpticalRecognitionofHandwritten & 42.2 $\pm$ 4.3 & 34 & 39.9 $\pm$ 2.7 & 37.6 & 42.8 $\pm$ 3.3 & 33.1 & 32.8 $\pm$ 1.7 & 48.6 & 36.4 $\pm$ 7.2 & 43 & 34.5 $\pm$ 3.1 & 46 &\textbf{ 25.8} $\pm$ 5.5 & 59.8 &\\[3pt]
 ConnectionistBenchData & 23.1 $\pm$ 3.7 & 61.4 & 21.5 $\pm$ 3.7 & 64.1 & 23.4 $\pm$ 3.0 & 60.8 & 18.2 $\pm$ 2.5 & 69.6 & 21.4 $\pm$ 2.7 & 64.2 & 20.2 $\pm$ 3.2 & 66.2 & \textbf{8.8} $\pm$ 4.9 & 85.4 \\[3pt]
 wdbc & 13.4 $\pm$ 3.6 & 55.1 & 13.4 $\pm$ 2.3 & 55.3 & 15.0 $\pm$ 3.4 & 50 &\textbf{ 9.9} $\pm$ 2.3 & 67 & 11.3 $\pm$ 2.1 & 62.1 & 11.3 $\pm$ 1.8 & 62.3 & 11.3 $\pm$ 4.2 & 62.5 \\[3pt]
 LungCancer & 17.0 $\pm$ 4.5 & 69.5 & 18.3 $\pm$ 3.6 & 67.2 & 17.2 $\pm$ 3.9 & 69.2 & 11.5 $\pm$ 2.5 & 79.4 & 19.6 $\pm$ 3.8 & 65 & 18.6 $\pm$ 4.0 & 66.7 & \textbf{5.8} $\pm$ 3.2 & 89.7 \\[3pt]
\hline
\end{tabular}
}
\end{table*}

\subsubsection{Results of Classification Accuracy on Training Datasets}
The classification accuracy results of the non-EC and EC algorithms on the training datasets are shown in Table 4 . In the upper half of Table 4, we observe that the SaWDE algorithm performs better than the other six non-EC methods in terms of classification accuracy across all twelve training datasets. The analysis of the data reveals the superiority of our algorithm. In fact, SaWDE improves the classification accuracy of more than half of the data by at least 5\%, some increasing by 15\%, compared to the best classification results of the six existing non-EC algorithms. Table 4 lower half shows that SaWDE also performs well when compared to the six EC algorithms, producing the best classification results in ten of the twelve datasets. The results show that our SaWDE algorithm performs effectively compared to existing EC and non-EC methods, providing a valid strategy for solving feature selection problems.

\begin{table*}[!htb]
\centering
\caption{Classification accuracy of six other non-EC, six EC methods and SaWDE on training dataSets}
\resizebox{125mm}{!}{
\begin{tabular}{l l l l l l l l }
\hline
 \multirow{2}{*}{Datasets} & \textbf{SFS} & \textbf{SBS} & \textbf{LRS21} &  \textbf{LRS32}  & \textbf{SFFS} & \textbf{SBFS}  & \textbf{SaWDE} \\
 \cline{2-8}
   \textbf{ } & \textbf{Mean $\pm$ Std} & \textbf{Mean $\pm$ Std}  & \textbf{Mean $\pm$ Std} & \textbf{Mean $\pm$ Std}  & \textbf{Mean $\pm$ Std} &  \textbf{Mean $\pm$ Std}  & \textbf{Mean $\pm$ Std} \\
\hline 
 grammatical$\_$facial$\_$expression01 & 0.8929  $\pm$ 0.0082 & 0.8147  $\pm$ 0.0113 & 0.8924  $\pm$ 0.0078 & 0.8965  $\pm$ 0.0077 & 0.9002  $\pm$ 0.0051 & 0.8170  $\pm$ 0.0108 & \textbf{0.9250} $\pm$ 0.0020 \\[3pt]
 SemeionHandwrittenDigit & 0.6832  $\pm$ 0.0492 & 0.8471  $\pm$ 0.0041 & 0.6772  $\pm$ 0.0511 & 0.6822  $\pm$ 0.0453 & 0.6396  $\pm$ 0.0350 & 0.8475  $\pm$ 0.0034 &  \textbf{0.9094} $\pm$ 0.0011 \\[3pt]
 isolet5 & 0.7481 $\pm$ 0.0392 & 0.7615  $\pm$ 0.0041 & 0.7460  $\pm$ 0.0277 & 0.7475  $\pm$ 0.0361 & 0.6949  $\pm$ 0.0363 & 0.7599  $\pm$ 0.0033 & \textbf{0.9100} $\pm$ 0.0041 \\[3pt]
 MultipleFeaturesDigit & 0.8449  $\pm$ 0.0208 & 0.9493  $\pm$ 0.0017 & 0.8453  $\pm$ 0.0209 & 0.8373  $\pm$ 0.0193 & 0.9194  $\pm$ 0.0150 & 0.9478  $\pm$ 0.0030 & \textbf{0.9843} $\pm$ 0.0000  \\[3pt]
 HAPTDataSet & 0.9244  $\pm$ 0.0111 & 0.9192  $\pm$ 0.0022 & 0.9261  $\pm$ 0.0081 & 0.9266  $\pm$ 0.0120 & 0.9282  $\pm$ 0.0074 & 0.9185  $\pm$ 0.0020 & \textbf{0.9702} $\pm$ 0.0000 \\[3pt]
 har & 0.9200  $\pm$ 0.0110 & 0.9134  $\pm$ 0.0031 & 0.9196  $\pm$ 0.0091 & 0.9170  $\pm$0.0114 & 0.9268  $\pm$ 0.0119 & 0.9137  $\pm$ 0.0031 & \textbf{0.9774} $\pm$ 0.0040  \\[3pt]
 UJIIndoorLoc & \textbf{1.0000}  $\pm$ 0.0000 & \textbf{1.0000}  $\pm$ 0.0000 & \textbf{1.0000}  $\pm$ 0.0000 & \textbf{1.0000}  $\pm$ 0.0000 & \textbf{1.0000}  $\pm$ 0.0000 & \textbf{1.0000}  $\pm$ 0.0000 & \textbf{1.0000}  $\pm$ 0.0000\\[3pt]
 MadelonValid & 0.6595 $\pm$ 0.0858 & 0.7155 $\pm$ 0.0072 & 0.7392 $\pm$ 0.1062 & 0.6803 $\pm$ 0.0804 & 0.8160 $\pm$ 0.0399 & 0.7145 $\pm$ 0.0060 & \textbf{0.9042} $\pm$ 0.0068  \\[3pt]
 OpticalRecognitionofHandwritten & 0.9482 $\pm$ 0.0108 & 0.9675 $\pm$ 0.0021 & 0.9490 $\pm$ 0.0116 & 0.9470 $\pm$ 0.0121 & 0.8185 $\pm$ 0.0737 & 0.9676 $\pm$ 0.0028  & \textbf{0.9857} $\pm$ 0.0020\\[3pt]
 ConnectionistBenchData & 0.8476 $\pm$ 0.0230 & 0.8095 $\pm$ 0.0093 & 0.8492 $\pm$ 0.0190 & 0.8444 $\pm$ 0.0181 & 0.8540 $\pm$ 0.0243 & 0.8080 $\pm$ 0.0111 &\textbf{ 0.9643} $\pm$ 0.0035  \\[3pt]
 wdbc & 0.9419 $\pm$ 0.0052 & 0.9474 $\pm$ 0.0023 & 0.9452 $\pm$ 0.0046 & 0.9449 $\pm$ 0.0051 & 0.9442 $\pm$ 0.0060 & 0.9466 $\pm$ 0.0024 & \textbf{0.9705} $\pm$ 0.0037 \\[3pt]
 LungCancer & 0.7300 $\pm$ 0.0657 & 0.6440 $\pm$ 0.0223 & 0.7167 $\pm$ 0.0460 & 0.7419 $\pm$ 0.0514 & 0.7470 $\pm$ 0.0481 & 0.6315 $\pm$ 0.0253 & \textbf{0.8929} $\pm$ 0.0275 \\[3pt]
\hline
\end{tabular}}

\centering
\resizebox{125mm}{!}{
\begin{tabular}{l l l l l l l l }
\hline
 \multirow{2}{*}{Datasets} & \textbf{GA} & \textbf{Original PSO} & \textbf{Standard PSO} &  \textbf{SaPSO}  & \textbf{DE} & \textbf{SaDE}  & \textbf{SaWDE} \\
 \cline{2-8}
   \textbf{ } & \textbf{Mean $\pm$ Std} & \textbf{Mean $\pm$ Std}  & \textbf{Mean $\pm$ Std} & \textbf{Mean $\pm$ Std}  & \textbf{Mean $\pm$ Std} &  \textbf{Mean $\pm$ Std}  & \textbf{Mean $\pm$ Std} \\
\hline 
grammatical$\_$facial$\_$expression01 & 0.9078  $\pm$ 0.0035 & 0.9107  $\pm$ 0.0022 & 0.9127  $\pm$ 0.0040 & 0.9159  $\pm$ 0.0026 & 0.8988  $\pm$ 0.0024 & 0.9029  $\pm$ 0.0023 &\textbf{0.9250 }$\pm$ 0.0020  \\[3pt]
 SemeionHandwrittenDigit & 0.8702  $\pm$ 0.0057 & 0.8753  $\pm$ 0.0041 & 0.8809  $\pm$ 0.0060 & 0.9031  $\pm$ 0.0049 & 0.8464  $\pm$ 0.0034 & 0.8510  $\pm$ 0.0044 & \textbf{0.9094} $\pm$ 0.0011  \\[3pt]
 isolet5 & 0.8205 $\pm$ 0.0076 & 0.8446  $\pm$ 0.0089 & 0.8495  $\pm$ 0.0138 & 0.8899  $\pm$ 0.0025 & 0.7941  $\pm$ 0.0035 & 0.8059  $\pm$ 0.0047 & \textbf{0.9100} $\pm$ 0.0041 \\[3pt]
 MultipleFeaturesDigit & 0.9688  $\pm$ 0.0029 & 0.9705  $\pm$ 0.0034 & 0.9734  $\pm$ 0.0041 & 0.9831  $\pm$ 0.0015 & 0.9585  $\pm$ 0.0017 & 0.9610  $\pm$ 0.0017 & \textbf{0.9843} $\pm$ 0.0000  \\[3pt]
 HAPTDataSet & 0.9467  $\pm$ 0.0032 & 0.9510  $\pm$ 0.0027 & 0.9529  $\pm$ 0.0024 & 0.9693  $\pm$ 0.0043 & 0.9351  $\pm$ 0.0020 & 0.9391  $\pm$ 0.0020 & \textbf{0.9702} $\pm$ 0.0000 \\[3pt]
 har & 0.9379  $\pm$ 0.0056 & 0.9429  $\pm$ 0.0056 & 0.9476  $\pm$ 0.0064 & 0.9745  $\pm$ 0.0048 & 0.9262  $\pm$ 0.0028 & 0.9269  $\pm$ 0.0025 & \textbf{0.9774} $\pm$ 0.0040  \\[3pt]
 UJIIndoorLoc & \textbf{1.0000}  $\pm$ 0.0000 & \textbf{1.0000}  $\pm$ 0.0000 & \textbf{1.0000}  $\pm$ 0.0000 & \textbf{1.0000}  $\pm$ 0.0000 & \textbf{1.0000}  $\pm$ 0.0000 & \textbf{1.0000}  $\pm$ 0.0000 & \textbf{1.0000}  $\pm$ 0.0000\\[3pt]
 MadelonValid & 0.7837 $\pm$ 0.0116 & 0.8157 $\pm$ 0.0110 & 0.8221 $\pm$ 0.0169 & 0.8722 $\pm$ 0.0050 & 0.7608 $\pm$ 0.0082 & 0.7724 $\pm$ 0.0046 & \textbf{0.9042} $\pm$ 0.0068  \\[3pt]
 OpticalRecognitionofHandwritten & 0.9860 $\pm$ 0.0018 & 0.9838 $\pm$ 0.0018 & 0.9849 $\pm$ 0.0019 & \textbf{0.9863} $\pm$ 0.0009 & 0.9693 $\pm$ 0.0023 & 0.9748 $\pm$ 0.0016   & 0.9857 $\pm$ 0.0020\\[3pt]
 ConnectionistBenchData & 0.9447 $\pm$ 0.0108 & 0.9467 $\pm$ 0.0114 & 0.9461 $\pm$ 0.0146 & 0.9537 $\pm$ 0.0086 & 0.9024 $\pm$ 0.0055 & 0.9124 $\pm$ 0.0062 & \textbf{0.9643} $\pm$ 0.0035  \\[3pt]
 wdbc & 0.9657 $\pm$ 0.0035 & 0.9670 $\pm$ 0.0037 & 0.9665 $\pm$ 0.0042 & 0.9682 $\pm$ 0.0033 & 0.9644 $\pm$ 0.0013 & 0.9658 $\pm$ 0.0014 &\textbf{ 0.9705} $\pm$ 0.0037 \\[3pt]
 LungCancer & 0.9355 $\pm$ 0.0289 & 0.9431 $\pm$ 0.0232 & \textbf{0.9484} $\pm$ 0.0303 & 0.9310 $\pm$ 0.0192 & 0.8718 $\pm$ 0.0245 & 0.8855 $\pm$ 0.0211 & 0.8929 $\pm$ 0.0275 \\[3pt]
\hline
\end{tabular}}
\end{table*}

\begin{figure*}[!htb]
\centering
\includegraphics[scale = 0.9]{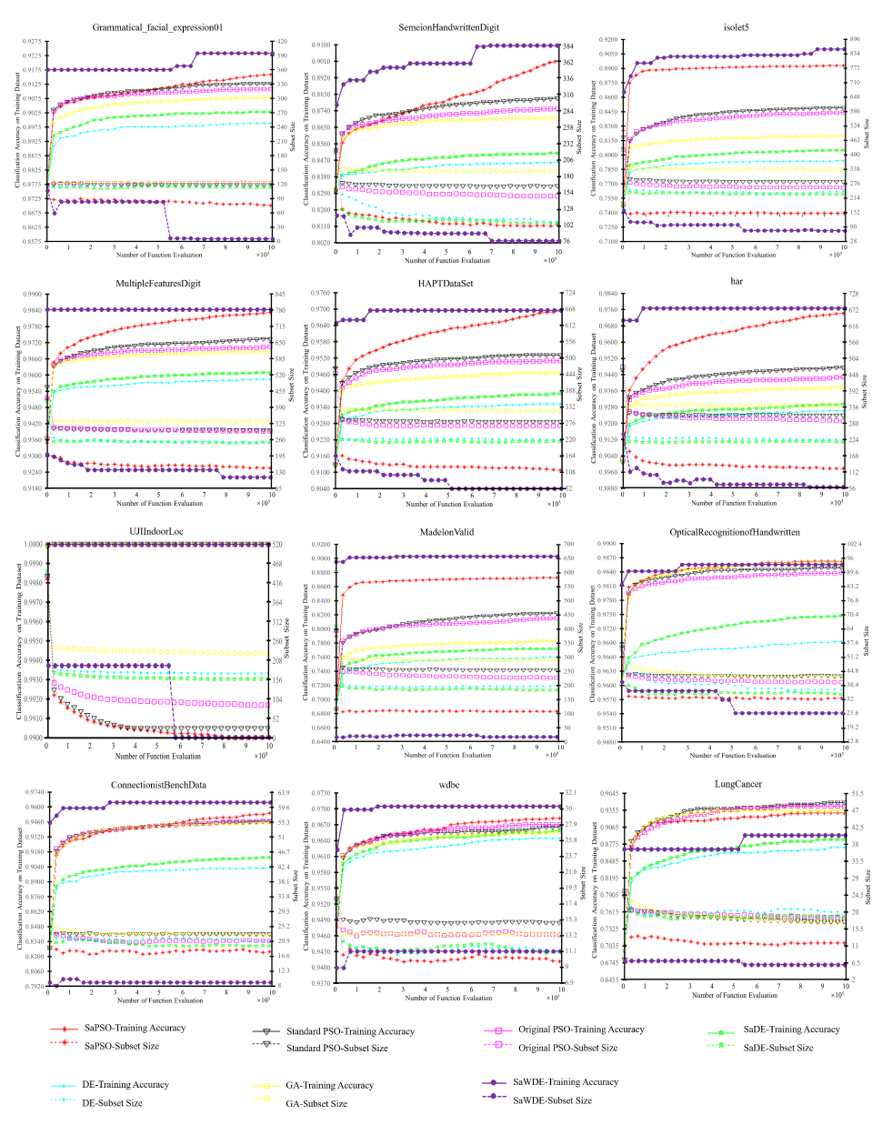}
\caption{ Comparative convergence curves of different algorithms in terms of training classification accuracy and feature subset size on different datasets. }
\label{fig:graph}
\end{figure*}

\subsubsection{Results of Convergence curves on Training Datasets}
The convergence performance comparison of SaWDE and other evolutionary methods in terms of the training classification accuracy and feature subset size are shown in Fig. 5. From this figure, it can be observed that all algorithms have similar convergence in the early stages. Specifically, SaWDE outperforms GA, original PSO, standard PSO, DE, SaDE, and SaPSO in terms of classification accuracy and feature subset size on most datasets even in the early convergence stage. Moreover, we can find that SaWDE still has excellent search ability in the late evolutionary stage. Therefore, based on the convergence curves, we can conclude that SaWDE is robust compared to other evolutionary models.

\subsubsection{Results of Classification Accuracy on Test Datasets}
Table 5 shows the classification accuracy results of non-EC and EC algorithms on the test datasets. The performance of the classification accuracy on the test datasets is an important measure of the robustness of an algorithm. As can be seen from Table 5, the classification accuracy of the SaWDE algorithm on test datasets perform well. Compared with the non-EC algorithms, the SaWDE algorithm performs better on all datasets except two. Compared to the EC algorithm, only one classification accuracy result is inferior to EC results. This shows that the SaWDE algorithm has better robustness and performs better than other algorithms in most cases.
\begin{table*}[!htb]
\centering
\caption{Classification accuracy of six other non-EC, six EC methods and SaWDE on test dataSets}
\resizebox{125mm}{!}{
\begin{tabular}{l l l l l l l l }
\hline
 \multirow{2}{*}{Datasets} & \textbf{SFS} & \textbf{SBS} & \textbf{LRS21} &  \textbf{LRS32}  & \textbf{SFFS} & \textbf{SBFS}  & \textbf{SaWDE} \\
 \cline{2-8}
   \textbf{ } & \textbf{Mean $\pm$ Std} & \textbf{Mean $\pm$ Std}  & \textbf{Mean $\pm$ Std} & \textbf{Mean $\pm$ Std}  & \textbf{Mean $\pm$ Std} &  \textbf{Mean $\pm$ Std}  & \textbf{Mean $\pm$ Std} \\
\hline 
 grammatical$\_$facial$\_$expression01 & 0.8447  $\pm$ 0.0133 & 0.7301  $\pm$ 0.0313 & 0.8471  $\pm$ 0.0150 & 0.8446  $\pm$ 0.0169 & 0.8165  $\pm$ 0.0211 & 0.7319  $\pm$ 0.0333 & \textbf{0.8967}$\pm$ 0.0059  \\[3pt]
 SemeionHandwrittenDigit & 0.5478  $\pm$ 0.0663 & \textbf{0.7994}  $\pm$ 0.0167 & 0.5495  $\pm$ 0.0613 & 0.5542  $\pm$ 0.0587 & 0.4355  $\pm$ 0.0548 & 0.7991  $\pm$ 0.0188 & 0.7963 $\pm$ 0.0062  \\[3pt]
 isolet5 & 0.6568 $\pm$ 0.0498 & 0.7144  $\pm$ 0.0162 & 0.6562  $\pm$ 0.0482 & 0.6586  $\pm$ 0.0505 & 0.4620  $\pm$ 0.0578 & 0.7139  $\pm$ 0.0167 & \textbf{0.8414} $\pm$ 0.0153 \\[3pt]
 MultipleFeaturesDigit & 0.7666  $\pm$ 0.0261 & 0.9308  $\pm$ 0.0090 & 0.7598  $\pm$ 0.0242 & 0.7609  $\pm$ 0.0250 & 0.8444  $\pm$ 0.0374 & 0.9308  $\pm$ 0.0087 & \textbf{0.9615} $\pm$ 0.0039  \\[3pt]
 HAPTDataSet & 0.8520  $\pm$ 0.0220 & 0.8479  $\pm$ 0.0130 & 0.8531  $\pm$ 0.0185 & 0.8516  $\pm$ 0.0219 & 0.7764  $\pm$ 0.0253 & 0.8449  $\pm$ 0.0127 & \textbf{0.9120} $\pm$ 0.0112 \\[3pt]
 har & 0.8702  $\pm$ 0.0204 & 0.8545  $\pm$ 0.0144 & 0.8779  $\pm$ 0.0165 & 0.8720  $\pm$ 0.0267 & 0.8017  $\pm$ 0.0397 & 0.8554  $\pm$ 0.0162 & \textbf{0.9271} $\pm$ 0.0098  \\[3pt]
 UJIIndoorLoc & 0.9981  $\pm$ 0.0032 & \textbf{0.9996}  $\pm$ 0.0011 & 0.9974  $\pm$ 0.0038 & 0.9989  $\pm$ 0.0021 & 0.9979  $\pm$ 0.0029 & 0.9993  $\pm$ 0.0022 & \textbf{0.9996}  $\pm$ 0.0001\\[3pt]
 MadelonValid & 0.5400 $\pm$ 0.0728 & 0.6173 $\pm$ 0.0291 & 0.6029 $\pm$ 0.1033 & 0.5797 $\pm$ 0.0779 & 0.6482 $\pm$ 0.0673 & 0.6141 $\pm$ 0.0325 & \textbf{0.7841} $\pm$ 0.0074  \\[3pt]
 OpticalRecognitionofHandwritten & 0.8833 $\pm$ 0.0216 & 0.9317 $\pm$ 0.0084 & 0.8843 $\pm$ 0.0244 & 0.8831 $\pm$ 0.0287 & 0.7248 $\pm$ 0.0789 & 0.9319 $\pm$ 0.0111  & \textbf{0.9493} $\pm$ 0.0037\\[3pt]
 ConnectionistBenchData & 0.6634 $\pm$ 0.0498 & 0.6513 $\pm$ 0.0624 & 0.6608 $\pm$ 0.0581 & 0.6456 $\pm$ 0.0391 & 0.6649 $\pm$ 0.0737 & 0.6494 $\pm$ 0.0595 & \textbf{0.7728} $\pm$ 0.0149  \\[3pt]
 wdbc & 0.8622 $\pm$ 0.0248 & 0.8887 $\pm$ 0.0148 & 0.8774 $\pm$ 0.0293 & 0.8719 $\pm$ 0.0300 & 0.8881 $\pm$ 0.0513 & 0.8881 $\pm$ 0.0141 & \textbf{0.9144} $\pm$ 0.0039 \\[3pt]
 LungCancer & 0.4870 $\pm$ 0.2037 & \textbf{0.6204} $\pm$ 0.0990 & 0.4417 $\pm$ 0.1694 & 0.4139 $\pm$ 0.1892 & 0.4269 $\pm$ 0.1658 & 0.5926 $\pm$ 0.1085 & 0.5455 $\pm$ 0.0134 \\[3pt]
\hline
\end{tabular}}

\centering
\resizebox{125mm}{!}{
\begin{tabular}{l l l l l l l l }
\hline
 \multirow{2}{*}{Datasets} & \textbf{GA} & \textbf{Original PSO} & \textbf{Standard PSO} & \textbf{SaPSO} &  \textbf{DE}  & \textbf{SaDE}  & \textbf{SaWDE} \\
 \cline{2-8}
   \textbf{ } & \textbf{Mean $\pm$ Std} & \textbf{Mean $\pm$ Std}  & \textbf{Mean $\pm$ Std} & \textbf{Mean $\pm$ Std}  & \textbf{Mean $\pm$ Std} &  \textbf{Mean $\pm$ Std}  & \textbf{Mean $\pm$ Std} \\
\hline 
 grammatical$\_$facial$\_$expression01 & 0.8513  $\pm$ 0.0151 & 0.8449  $\pm$ 0.0131 & 0.8507  $\pm$ 0.0126 & 0.8448  $\pm$ 0.0157 & 0.8506  $\pm$ 0.0133 & 0.8470  $\pm$ 0.0117 & \textbf{0.8967} $\pm$ 0.0059  \\[3pt]
 SemeionHandwrittenDigit & 0.7872  $\pm$ 0.0201 & 0.7821  $\pm$ 0.0272 & 0.7892  $\pm$ 0.0254 & 0.7795  $\pm$ 0.0216 & 0.7690  $\pm$ 0.0207 & 0.7685  $\pm$ 0.0183 & \textbf{0.7963} $\pm$ 0.0062  \\[3pt]
 isolet5 & 0.7487 $\pm$ 0.0258 & 0.7875  $\pm$ 0.0232 & 0.7777  $\pm$ 0.0202 & 0.8106  $\pm$ 0.0167 & 0.7375  $\pm$ 0.0214 & 0.7534  $\pm$ 0.0223  & \textbf{0.8414} $\pm$ 0.0153 \\[3pt]
 MultipleFeaturesDigit & 0.9356  $\pm$ 0.0091 & 0.9327  $\pm$ 0.0105 & 0.9348  $\pm$ 0.0125 & 0.9388  $\pm$ 0.0092 & 0.9300  $\pm$ 0.0106 & 0.9334  $\pm$ 0.0104 & \textbf{0.9615} $\pm$ 0.0039  \\[3pt]
 HAPTDataSet & 0.8530  $\pm$ 0.0159 & 0.8585  $\pm$ 0.0146 & 0.8599  $\pm$ 0.0152 & 0.8814  $\pm$ 0.0121 & 0.8496  $\pm$ 0.0199 & 0.8504  $\pm$ 0.0172 & \textbf{0.9120} $\pm$ 0.0112 \\[3pt]
 har & 0.8679  $\pm$ 0.0146 & 0.8727  $\pm$ 0.0156 & 0.8713  $\pm$ 0.0174 & 0.9140  $\pm$ 0.0212 & 0.8630  $\pm$ 0.0212 & 0.8673  $\pm$ 0.0243 & \textbf{0.9271} $\pm$ 0.0098  \\[3pt]
 UJIIndoorLoc & 0.9962  $\pm$ 0.0053 & 0.9901  $\pm$ 0.0064 & 0.9952  $\pm$ 0.0043 & 0.9986  $\pm$ 0.0026 & 0.9908  $\pm$ 0.0082 & 0.9919  $\pm$ 0.0103 & \textbf{0.9996}  $\pm$ 0.0001\\[3pt]
 MadelonValid & 0.6421 $\pm$ 0.0291 & 0.6598 $\pm$ 0.0352 & 0.6503 $\pm$ 0.0402 & 0.6863 $\pm$ 0.0351 & 0.6400 $\pm$ 0.0357 & 0.6387 $\pm$ 0.0346 & \textbf{0.7841} $\pm$ 0.0074  \\[3pt]
 OpticalRecognitionofHandwritten & 0.9312 $\pm$ 0.0097 & 0.9278 $\pm$ 0.0098 & 0.9324 $\pm$ 0.0078 & 0.9364 $\pm$ 0.0107 & 0.9187 $\pm$ 0.0135 & 0.9247 $\pm$ 0.0136  & \textbf{0.9493} $\pm$ 0.0037\\[3pt]
 ConnectionistBenchData & 0.6908 $\pm$ 0.0556 & 0.6972 $\pm$ 0.0462 & 0.6804 $\pm$ 0.0423 & 0.7005 $\pm$ 0.0511 & 0.6779 $\pm$ 0.0442 & 0.6841 $\pm$ 0.0553 & \textbf{0.7728} $\pm$ 0.0149  \\[3pt]
 wdbc & 0.9125 $\pm$ 0.0232 & \textbf{0.9162} $\pm$ 0.0191 & 0.9092 $\pm$ 0.0213 & 0.9099 $\pm$ 0.0135 & 0.9021 $\pm$ 0.0200 & 0.9021 $\pm$ 0.0127 & 0.9144 $\pm$ 0.0039 \\[3pt]
 LungCancer & 0.4750 $\pm$ 0.1602 & 0.4565 $\pm$ 0.1211 & 0.4565 $\pm$ 0.1348 & 0.5102 $\pm$ 0.1442 & 0.4852 $\pm$ 0.1523 & 0.4481 $\pm$ 0.1446 & \textbf{0.5455} $\pm$ 0.0134 \\[3pt]
\hline
\end{tabular}}
\end{table*}

\subsection{Performance of Different Mutation Strategies}
Fig. 6 shows the mean classification accuracy of all eight \emph{CMSs} and our SaWDE algorithm on training and test datasets. The heatmap $(a)$ represents the classification accuracy results on the training datasets, the red indicating the superior results. The diagram $(b)$ shows the classification accuracy results on the test datasets. The different datasets are denoted by specific colors, and the best result in each dataset has a different color.    

\begin{figure*}[!htb]
\centering
\includegraphics[scale = 0.37]{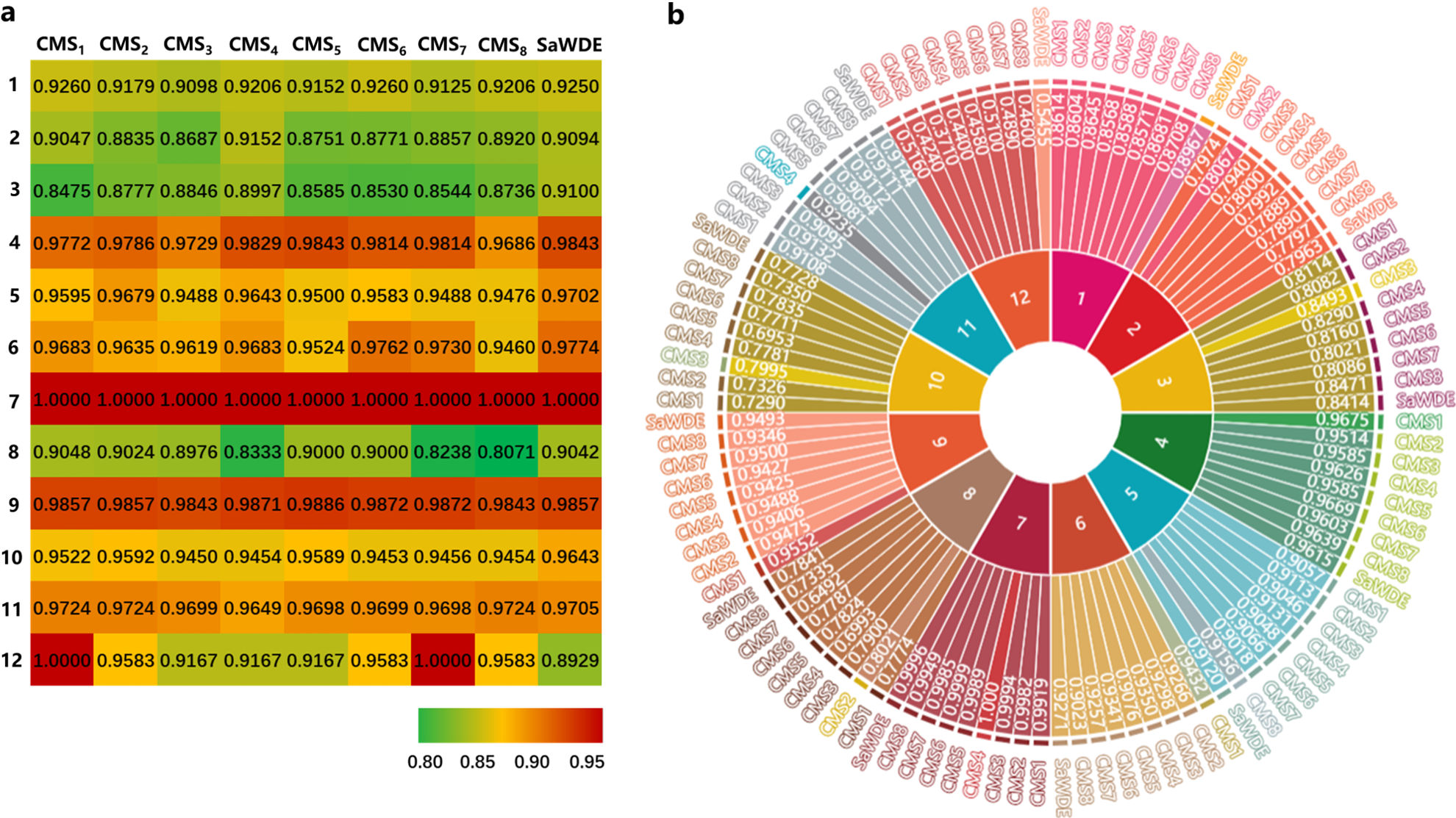}
\caption{\raggedright The subset size on training datasets (a) and test datasets (b). The numbers 1 to 12 in both (a) and (b) denote the data grammatical$\_$facial$\_$expression01, SemeionHandwrittenDigit, isolet5, MultipleFeaturesDigit, HAPTDataSet, har, UJIIndoorLoc, MadelonValid, OpticalRecognitionofHandwritten, ConnectionistBenchData, wdbc and LungCancer, respectively.}
\label{fig:graph}
\end{figure*}

\begin{figure*}[!htb]
\centering
\includegraphics[scale = 0.39]{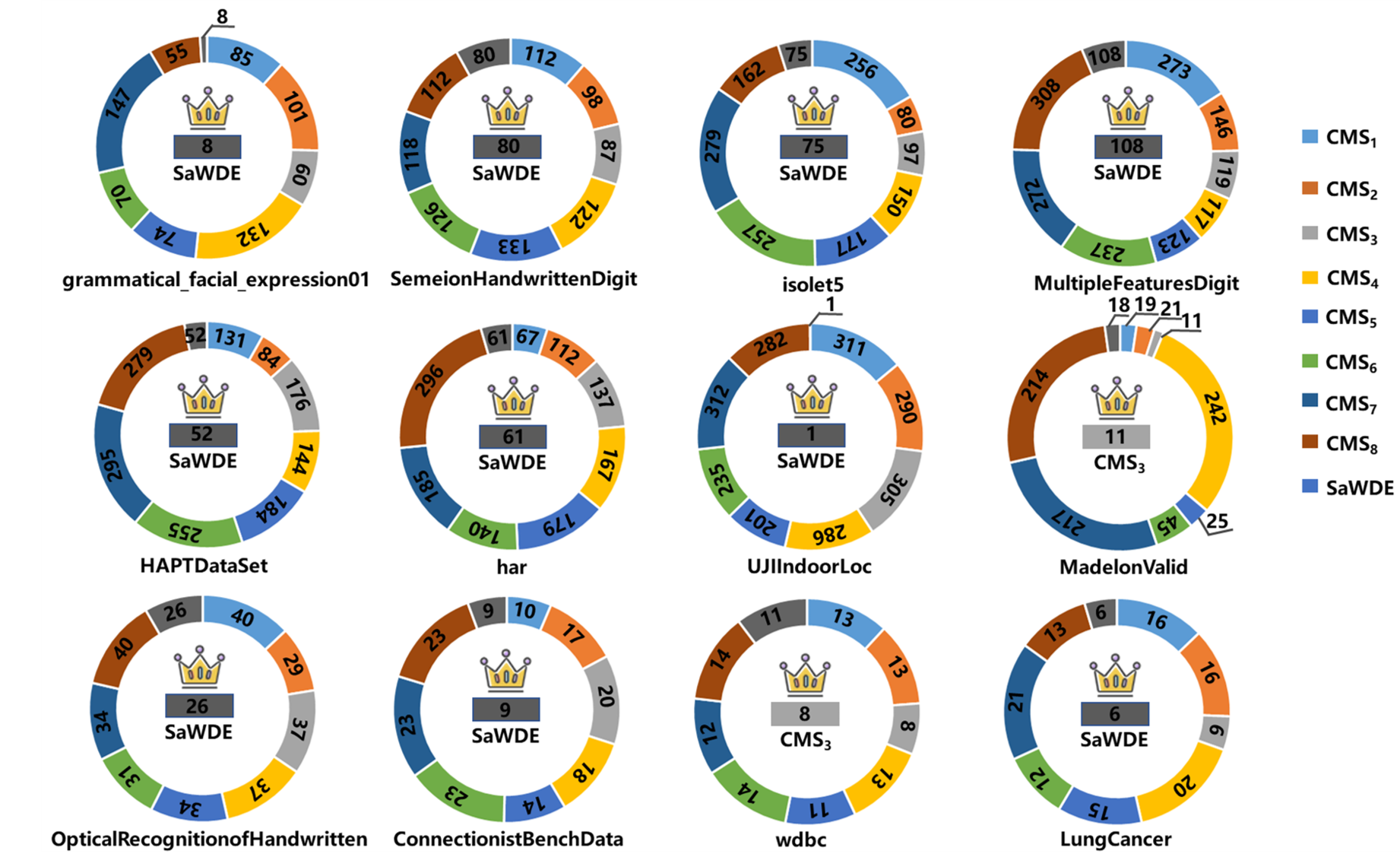}
\caption{ The classification accuracy on each single \emph{CMS} and SaWDE on training datasets. The numbers on the ring denote the corresponding algorithms' subset size. The  best result of all of the algorithms is shown beneath the crown in the center of each ring. The size is rounded off here.}
\label{fig:graph}
\end{figure*}

Fig. 7 shows the mean subset size of all eight \emph{CMSs} and our SaWDE algorithm on training datasets. In Fig. 7, each color on the ring represents a different algorithm, and the number in the color denotes the subset size. The best result of all of the algorithms is shown beneath the crown in the center of each ring. As can be seen from Fig. 7, SaWDE generally delivers the best results in terms of the size of the solution.
 
In the dataset, grammatical$\_$facial$\_$expression01, the three with the worst accuracy for classification are \emph{CMS}$_3$, \emph{CMS}$_5$ and \emph{CMS}$_7$. The worst three in the data, SemeionHandwrittenDigit are \emph{CMS}$_3$, \emph{CMS}$_5$, and \emph{CMS}$_6$; in the data, isolet5 \emph{CMS}$_1$, \emph{CMS}$_6$ and \emph{CMS}$_7$; in the data, MultipleFeaturesDigit \emph{CMS}$_1$, \emph{CMS}$_3$, and \emph{CMS}$_8$; in the data, HAPTDataSet \emph{CMS}$_3$, \emph{CMS}$_7$, and \emph{CMS}$_8$ and in the data, 
har \emph{CMS}$_3$, \emph{CMS}$_5$, and \emph{CMS}$_8$. The classification accuracy of all strategies in UJIIndoorLoc is the same, so it is not counted. The worst three in the data, MadelonValid are \emph{CMS}$_4$, \emph{CMS}$_7$ and \emph{CMS}$_8$. Data OpticalRecognitionofHandwritten worst values contain four \emph{CMSs}, therefore, statistics of four \emph{CMS}$_1$, \emph{CMS}$_2$, \emph{CMS}$_3$ and \emph{CMS}$_8$ respectively. The worst three in the data, ConnectionistBenchData are \emph{CMS}$_3$, \emph{CMS}$_6$, and \emph{CMS}$_8$; in data WDBC, are \emph{CMS}$_4$, \emph{CMS}$_5$, and \emph{CMS}$_7$; and data LungCancer, \emph{CMS}$_3$, \emph{CMS}$_4$ and \emph{CMS}$_5$. Among these, \emph{CMS}$_1$, \emph{CMS}$_2$, \emph{CMS}$_3$, \emph{CMS}$_4$, \emph{CMS}$_5$, \emph{CMS}$_6$, \emph{CMS}$_7$ and \emph{CMS}$_8$ are counted 3 times, 1 time, 8 times, 3 times, 5 times, 3 times, 5 times and 6 times, respectively. Therefore, \emph{CMS}$_3$ with 8 counts and \emph{CMS}$_8$ with 6 counts are excluded first. \emph{CMS}$_5$ and \emph{CMS}$_7$, were counted 5 times, but \emph{CMS}$_7$ is eliminated because only 3 datasets of \emph{CMS}$_5$ are not as accurate as \emph{CMS}$_7$. Based on the above, \emph{CMS}$_1$, \emph{CMS}$_2$, \emph{CMS}$_4$, \emph{CMS}$_5$, and \emph{CMS}$_6$ are selected for further strategy pool building.

\begin{figure}[!htbp]
\centering
\includegraphics[width=8.5cm]{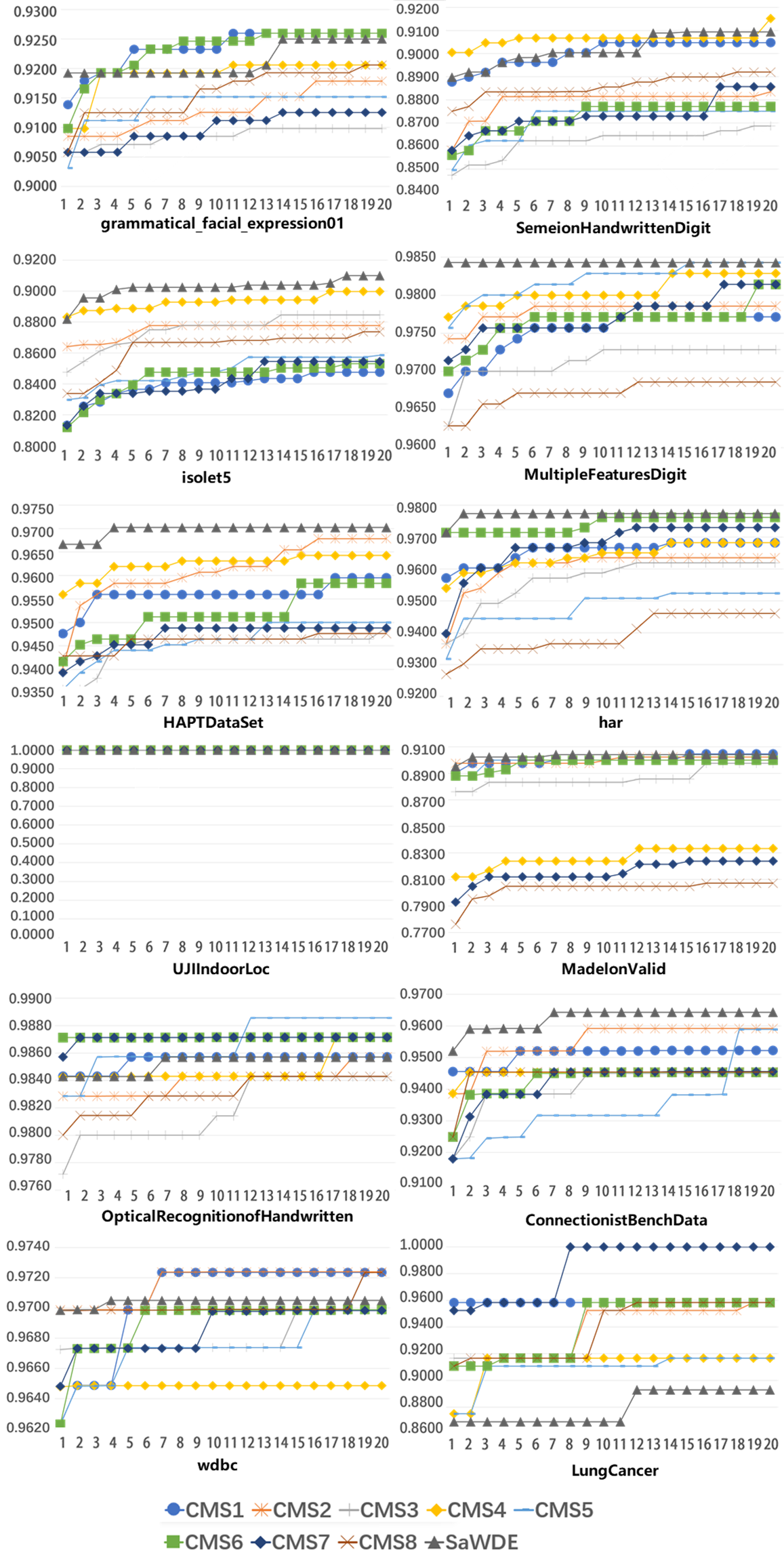}
\quad
\caption{ Classification accuracy convergence curves on data 1-data 12. Each shape denotes an algorithm, and the numbers in abscissa denote the function evaluation times, which the unit of measurement is $5 \times 10^4$ FES.}
\end{figure}

Fig. 8 depicts the convergence curve of classification accuracy from \emph{CMS}$_1$ to \emph{CMS}$_8$ and SaWDE on all datasets, respectively. It can be seen from the results that different strategies have varying degrees of convergence at the different stages. Given that the search capability of each strategy is different, the combination of strategies and the construction of a strategy pool can efficiently extend the search capability of the algorithm and prevent it falling into the local optimal situation. From the performance of the SaWDE algorithm on all datasets, it compares to the other eight separate \emph{CMSs}, getting the best results on most datasets.

\subsection{Influence of Different Population Sizes on Datasets}
Fig. 9 represents the results of our investigation on the subset size on the training datasets. Figs. 10 - 11 represent the results of our investigation on the classification accuracy of different population sizes on the training datasets and the classification accuracy on the test datasets respectively. The aim of this experiment is to find the most suitable population number to get the best evolutionary results.

In this section, we compare our SaWDE algorithm with SaPSO to optimize the feature subsets. As shown in Fig. 9, when $N$ is 50 or 500, all the subset sizes of SaWDE are smaller than those of SaPSO. When $N$ is 100, 200, or 300, there are one, three, and two slightly bigger solutions than SaPSO, respectively. Different populations can have some effect on subset size, but overall, SaWDE has an advantage over SaPSO in terms of subset size.

We see that when $N$ is 50 in Fig. 10, SaWDE has only five results that are better than or equal to SaPSO's results. This indicates that when the population size is 50, there is a certain degree of loss of diversity. However, when $N$ is 100, 200, 300, or 500 respectively, SaWDE has only two results inferior to SaPSO, which indicates that when the population number is greater than or equal to 100, the diversity requirements of the population can be basically satisfied.

\begin{figure}[!htbp]
\centering
\includegraphics[width=8.05cm]{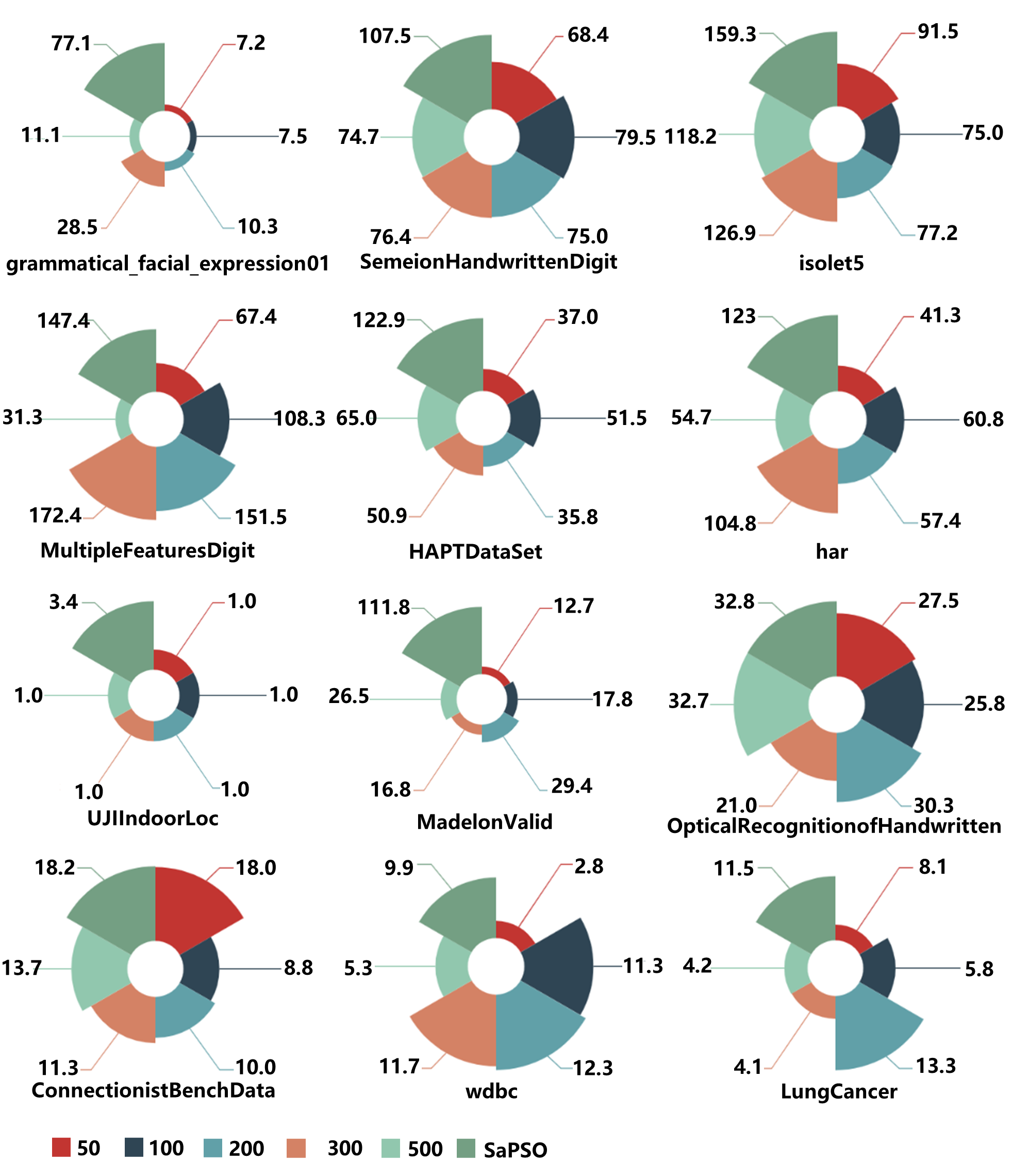}
\caption{The influence of different N on subset size. Each color denotes a situation, and the bigger the area, indicating the subset size is bigger.}
\end{figure}

We see that when $N$ is 50 or 300 in Fig. 11, SaWDE's classification accuracy on the test dataset is in one situation worse than SaPSO's, and when $N$ is 200 or 500, SaPSO's classification accuracy is in two situations. However, when $N$ is 100, SaWDE's classification accuracy on all test datasets is superior to SaPSO's. Combining the above three aspects and considering other problems such as computational complexity, we finally set the population to a size of 100.

\begin{figure}[!htbp]
\centering
\includegraphics[width=7.4cm]{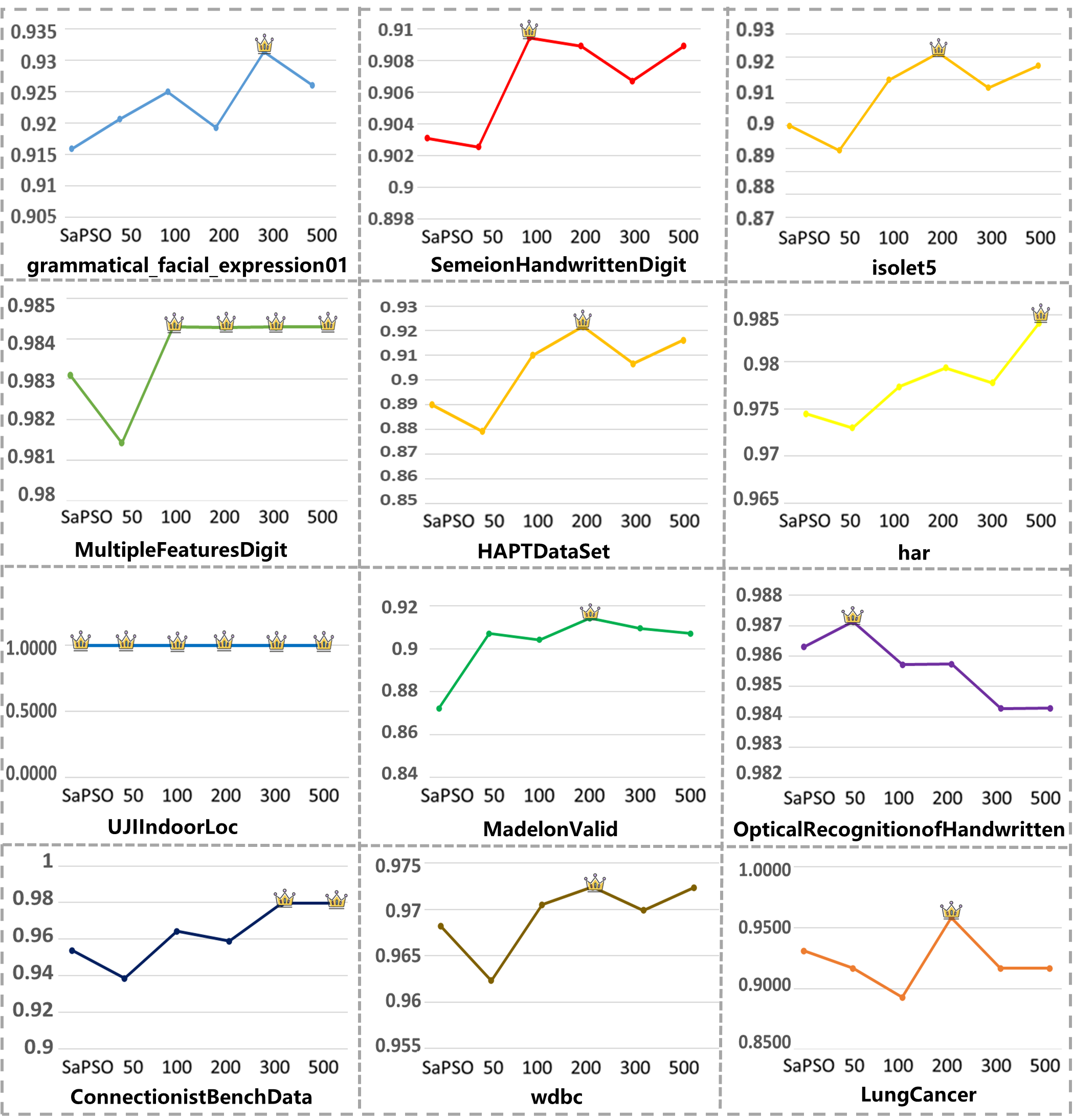}
\caption{ The influence of different N on classification accuracy of training  datasets. The classification accuracy of each case is connected with broken lines to show the difference more clearly, and there is a crown in the best case.}
\end{figure}

\begin{figure}[!htbp]
\centering
\includegraphics[width=7.4cm]{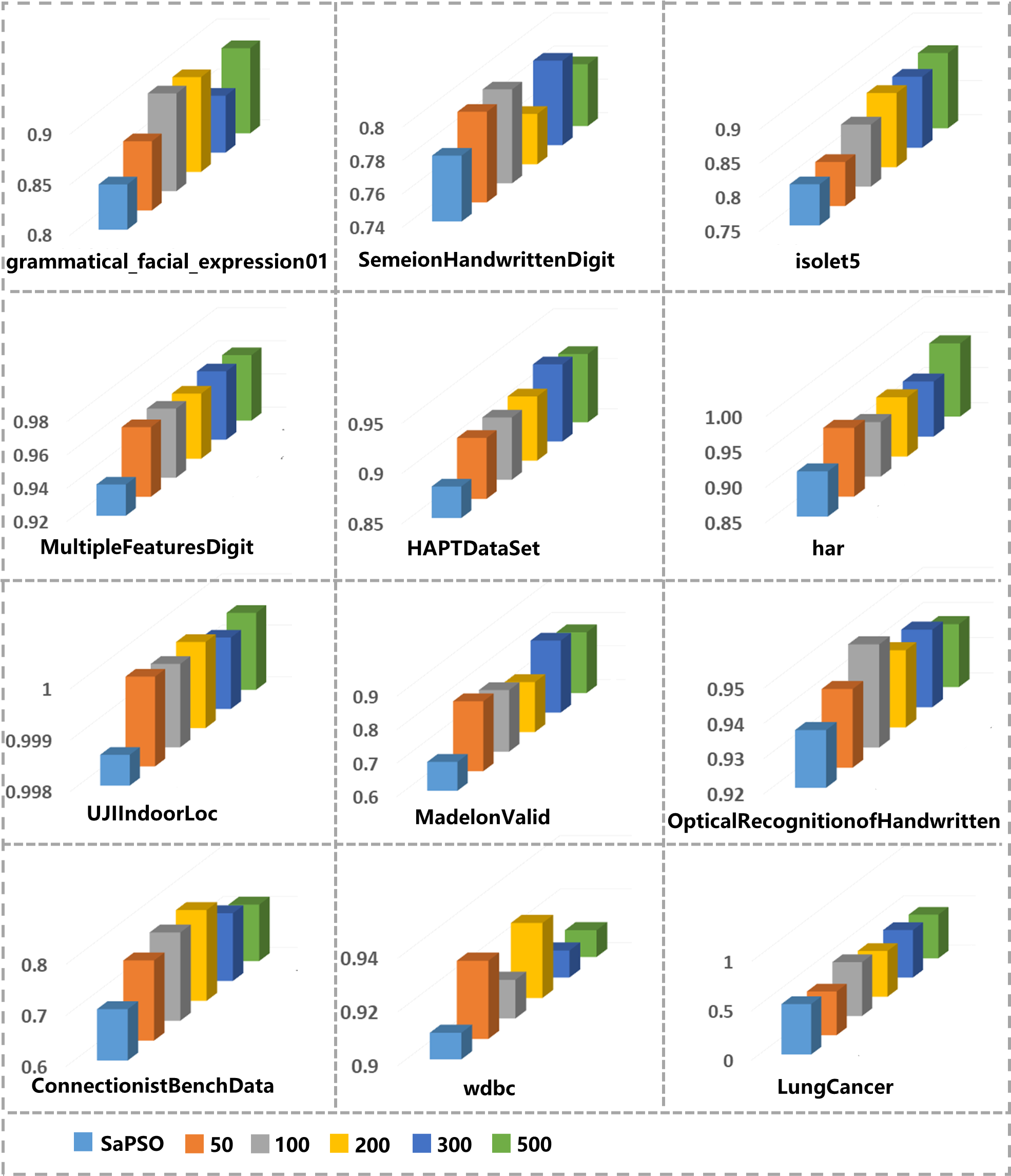}
\caption{\raggedright The influence of different N on classification accuracy of test datasets. The column is used to represent accuracy, which the higher the column indicates the better results.}
\end{figure}

\subsection{Effect of the Number of Different sub-populations on Datasets}

The number of sub-populations can affect the degree of evolution; too few populations may not achieve the desired results, while too many populations affect the complexity of the algorithm. To address this problem, we have conducted an experiment to discuss the effect of the number of sub-populations. In our study, since all the sub-populations are of the same size, we set the number of sub-populations at 2, 4, 5, and 10 to enable the population of sub-populations to be divisible by the number of total populations.

\begin{table*}[!htb]
\centering
\caption{Classification accuracy of different numbers of sub-populations on training datasets}
\resizebox{125mm}{!}{
\begin{tabular}{|l |l| l| l| l|}
\hline
\textbf{Datasets} & \textbf{2-Subs}  & \textbf{4-Subs} & \textbf{5-Subs} & \textbf{10-Subs}  \\\hline
grammatical$\_$facial$\_$expression01	&	0.9233	&	0.9246	&	\textbf{0.9250}	&	0.9179	\\\hline
SemeionHandwrittenDigit	&	0.9026	&	\textbf{0.9195}	&	0.9094	&	0.9048	\\\hline
isolet5	&	0.9080	&	0.8970	&	\textbf{0.9100}	&	0.8860	\\\hline
MultipleFeaturesDigit	&	\textbf{0.9843}	&	\textbf{0.9843}	&	\textbf{0.9843}	&	0.9829	\\\hline
HAPTDataSet	&	0.9679	&	\textbf{0.9714}	&	0.9702	&	0.9679	\\\hline
har	&	0.9746	&	0.9730	&	0.9774	&	\textbf{0.9810}	\\\hline
UJIIndoorLoc	&	\textbf{1.0000}	&	\textbf{1.0000}	&	\textbf{1.0000}	&	\textbf{1.0000}	\\\hline
MadelonValid	&	0.8952	&	\textbf{0.9592}	&	0.9042	&	0.9048	\\\hline
OpticalRecognitionofHandwritten	&	0.9800	&	0.9648	&	\textbf{0.9857}	&	0.9800	\\\hline
ConnectionistBenchData	&	0.9521	&	0.9167	&	\textbf{0.9643}	&	0.9386	\\\hline
wdbc	&	0.9674	&	0.8976	&	\textbf{0.9705}	&	0.9599	\\\hline
LungCancer	&	0.9167	&	\textbf{0.9843}	&	0.8929	&	0.8750	\\\hline
\textbf{Average}	&	0.9477	&	0.9494	&	\textbf{0.9495}	&	0.9415	\\
\hline
\end{tabular}}
\centering
\end{table*}

\begin{table*}[!htb]
\centering
\caption{Classification accuracy of different numbers of sub-populations on test datasets}
\resizebox{125mm}{!}{
\begin{tabular}{|l |l| l| l| l|}
\hline
\textbf{Datasets} & \textbf{2-Subs}  & \textbf{4-Subs} & \textbf{5-Subs} & \textbf{10-Subs}  \\\hline
grammatical$\_$facial$\_$expression01	&	0.8934	&	0.8903	&	\textbf{0.8967}	&	0.8903	\\	\hline
SemeionHandwrittenDigit	&	0.7980	&	\textbf{0.8128}	&	0.7963	&	0.8079	\\	\hline
isolet5	&	0.8205	&	0.8333	&	\textbf{0.8414}	&	0.8013	\\	\hline
MultipleFeaturesDigit	&	0.9633	&	\textbf{0.9700}	&	0.9615	&	0.9600	\\	\hline
HAPTDataSet	&	\textbf{0.9278}	&	0.8917	&	0.9120	&	0.8972	\\	\hline
har	&	0.9407	&	\textbf{0.9556}	&	0.9271	&	0.9185	\\	\hline
UJIIndoorLoc	&	\textbf{1.0000}	&	\textbf{1.0000}	&	0.9996	&	\textbf{1.0000}	\\	\hline
MadelonValid	&	0.7444	&	0.7419	&	\textbf{0.7841}	&	0.7778	\\	\hline
OpticalRecognitionofHandwritten	&	0.9233	&	0.9298	&	\textbf{0.9493}	&	0.9300	\\	\hline
ConnectionistBenchData	&	0.7742	&	0.3000	&	0.7728	&	\textbf{0.7903}	\\	\hline
wdbc	&	0.9006	&	0.8056	&	0.9144	&	\textbf{0.9415}	\\	\hline
LungCancer	&	0.2000	&	\textbf{0.9233}	&	0.5455	&	0.6000	\\	\hline
\textbf{Average}	&	0.8239	&	0.8379	&	0.8584	&	\textbf{0.8596}	\\	
\hline
\end{tabular}}
\centering
\end{table*}

The experimental results are summarized in Tables 6-7. The best results for each dataset are highlighted in bold text. Meanwhile, the average performance of each algorithm on all datasets are tabulated in the last row of each Table. In those tables, 2-Subs, 4-Subs, 5-Subs and 10-Subs denote the proposed algorithm with 2, 4, 5, and 10 sub-populations, respectively.

Table 6 summarized the classification accuracy of different numbers of sub-populations on training datasets. It can be observed that the algorithm with  the 5 sub-populations can achieves the best 7 results out of 12 on the training set. Besides, it can provide the best average performance. After that, we also provide the training models with different number of sub-populations to predict the test data. Table 7 tabulates the performance of the algorithm on test datasets with different numbers of sub-populations. We can find that the 10-Subs provided the best performance for the test data, while the 5-Subs gave similar results to the 10-Subs. However, 10-Subs cannot present better results in the training phase. According to hyperparameter optimization in machine learning \cite{bergstra2011algorithms}, we need to discuss the parameters in the training phase rather than the test phase. Therefore, the number of sub-populations was set to 5 to conduct a fair comparison.

\subsection{Extend Performance Comparisons for Higher-dimensional Datasets}
To illustrate the robustness and generalization ability of SaWDE, twelve higher-dimensional datasets are used for further experiments in this section. The experimental results are tabulated in Tables 8 - 10. The best results for each dataset are highlighted in bold text. Meanwhile, we set the average performance of each algorithm on all datasets in the last row of each table.

\begin{table*}[htb]
\centering
\caption{Test accuracy of six EC methods and SaWDE on twelve higher-dimensional datasets}
\resizebox{125mm}{!}{
\begin{tabular}{|l |l| l| l| l| l| l| l|}
\hline
\textbf{Datasets} & \textbf{GA}  & \textbf{Original PSO} & \textbf{Standard PSO} & \textbf{SaPSO}  & \textbf{DE}  & \textbf{SaDE} & \textbf{SaWDE} \\
\hline 
Alizadeh-2000-v1	&	0.6667 	&	\textbf{0.8333} 	&	0.5833 	&	0.5000 	&	\textbf{0.8333} 	&	0.7500 	&	\textbf{0.8333} 	\\	\hline
Alizadeh-2000-v2	&	0.8333 	&	\textbf{1.0000} 	&	0.9444 	&	0.8889 	&	0.9444 	&	\textbf{1.0000} 	&	\textbf{1.0000} 	\\	\hline
Armstrong-2002-v1	&	0.8571 	&	0.8571 	&	0.9048 	&	0.9107 	&	0.9048 	&	\textbf{0.9524} 	&	0.9048 	\\	\hline
Bittner-2000	&	0.7273 	&	0.6364 	&	0.7273 	&	0.7500 	&	\textbf{0.8182} 	&	0.7273 	&	0.7273 	\\	\hline
Dyrskjot-2003	&	0.6667 	&	0.7500 	&	0.7500 	&	0.5833 	&	0.6667 	&	0.7500 	&	\textbf{0.8333} 	\\	\hline
Garber-2001	&	0.6842 	&	\textbf{0.8421} 	&	\textbf{0.8421} 	&	0.8095 	&	0.7895 	&	0.7895 	&	\textbf{0.8421} 	\\	\hline
Liang-2005	&	\textbf{0.8182}	&	\textbf{0.8182} & \textbf{0.8182} &	0.7222 	&	\textbf{0.8182}	&	\textbf{0.8182}	&	\textbf{0.8182} 	\\	\hline
Nutt-2003-v2	&	0.6250 	&	0.6250 	&	0.6250 	&	0.5000 	&	0.5000 	&	\textbf{0.8750} 	&	\textbf{0.8750} 	\\	\hline
Pomeroy-2002-v1	&	\textbf{0.9000} 	&	0.7000 	&	0.7000 	&	0.6111 	&	0.6000 	&	\textbf{0.9000} 	&	0.8000 	\\	\hline
Pomeroy-2002-v2	&	\textbf{0.7500} 	&	0.5833 	&	0.6667 	&	0.5222 	&	0.5833 	&	\textbf{0.7500} 	&	0.6667 	\\	\hline
Shipp-2002-v1	&	0.8261 	&	0.7826 	&	0.7391 	&	0.7798 	&	\textbf{0.9130} 	&	0.8261 	&	0.8696 	\\	\hline
West-2001	&	0.4286 	&	0.7143 	&	0.6429 	&	0.8500 	&	0.7143 	&	0.7857 	&	\textbf{0.8571} 	\\	\hline
\textbf{Average}	&	0.7319 	&	0.7619 	&	0.7453 	&	0.7023 	&	0.7571 	&	0.8270 	&	\textbf{0.8356} 	\\	\hline

\end{tabular}}
\centering
\end{table*}

\begin{table*}[htb]
\centering
\caption{Subset size of six EC methods and SaWDE on twelve higher-dimensional datasets}
\resizebox{125mm}{!}{
\begin{tabular}{|l |l| l| l| l| l| l| l|}
\hline
\textbf{Datasets} & \textbf{GA}  & \textbf{Original PSO} & \textbf{Standard PSO} & \textbf{SaPSO}  & \textbf{DE}  & \textbf{SaDE} & \textbf{SaWDE} \\
\hline 
Alizadeh-2000-v1	&	564.0 	&	505.0 	&	557.0 	&	\textbf{9.0} 	&	563.0 	&	548.0 	&	217.0 	\\	\hline
Alizadeh-2000-v2	&	1059.0 	&	1036.0 	&	1016.0 	&	\textbf{23.0} 	&	1037.0 	&	1044.0 	&	833.0 	\\	\hline
Armstrong-2002-v1	&	547.0 	&	526.0 	&	550.0 	&	\textbf{11.0} 	&	537.0 	&	507.0 	&	411.0 	\\	\hline
Bittner-2000	&	1160.0 	&	1088.0 	&	1114.0 	&	\textbf{29.0} 	&	1081.0 	&	1091.0 	&	110.0 	\\	\hline
Dyrskjot-2003	&	598.0 	&	597.0 	&	583.0 	&	\textbf{6.0} 	&	595.0 	&	593.0 	&	401.0 	\\	\hline
Garber-2001	&	2253.0 	&	2240.0 	&	2256.0 	&	1173.0 	&	2956.0 	&	2278.0 	&	\textbf{3.0} 	\\	\hline
Liang-2005	&	673.0 	&	708.0 	&	710.0 	&	4.0 	&	712.0 	&	698.0 	&	\textbf{1.0} 	\\	\hline
Nutt-2003-v2	&	526.0 	&	526.0 	&	548.0 	&	\textbf{5.0} 	&	558.0 	&	542.0 	&	191.0 	\\	\hline
Pomeroy-2002-v1	&	447.0 	&	440.0 	&	437.0 	&	\textbf{6.0} 	&	437.0 	&	420.0 	&	130.0 	\\	\hline
Pomeroy-2002-v2	&	680.0 	&	660.0 	&	666.0 	&	282.0 	&	804.0 	&	657.0 	&	\textbf{5.0} 	\\	\hline
Shipp-2002-v1	&	385.0 	&	376.0 	&	387.0 	&	\textbf{103.0} 	&	387.0 	&	407.0 	&	243.0 	\\	\hline
West-2001	&	589.0 	&	598.0 	&	609.0 	&	\textbf{9.0} 	&	616.0 	&	583.0 	&	245.0 	\\	\hline
\textbf{Average}	&	790.1 	&	775.0 	&	786.1 	&	\textbf{138.3} 	&	856.9 	&	780.7 	&	232.5 	\\	\hline

\end{tabular}}
\end{table*}

\begin{table*}[htb!]
\centering
\caption{CPU time of six EC methods and SaWDE on twelve higher-dimensional datasets}
\centering
\resizebox{125mm}{!}{
\begin{tabular}{|l |l| l| l| l| l| l| l|}
\hline
\textbf{Datasets} & \textbf{GA}  & \textbf{Original PSO} & \textbf{Standard PSO} & \textbf{SaPSO}  & \textbf{DE}  & \textbf{SaDE} & \textbf{SaWDE} \\
\hline 
Alizadeh-2000-v1	&	9624.1 	&	10088.5 	&	10373.7 	&	9472.2 	&	10089.5 	&	9799.1 	&	\textbf{67.8} 	\\	\hline
Alizadeh-2000-v2	&	10227.5 	&	11332.6 	&	11703.9 	&	9460.7 	&	10966.4 	&	10542.3 	&	\textbf{15.9} 	\\	\hline
Armstrong-2002-v1	&	9677.1 	&	10604.9 	&	10276.6 	&	10496.1 	&	10140.4 	&	9939.2 	&	\textbf{6.8} 	\\	\hline
Bittner-2000	&	9933.9 	&	11154.0 	&	11264.5 	&	11153.6 	&	10505.9 	&	10084.0 	&	\textbf{16.2} 	\\	\hline
Dyrskjot-2003	&	9528.7 	&	10482.4 	&	10105.1 	&	10553.1 	&	9840.0 	&	9685.7 	&	\textbf{38.0} 	\\	\hline
Garber-2001	&	\textbf{11238.3} 	&	14121.2 	&	14264.2 	&	17146.3 	&	13098.3 	&	12787.3 	&	68932.2 	\\	\hline
Liang-2005	&	11544.8 	&	12094.2 	&	11767.9 	&	11005.1 	&	15487.9 	&	\textbf{9746.0} 	&	19040.3 	\\	\hline
Nutt-2003-v2	&	9467.6 	&	9978.9 	&	9914.4 	&	10594.2 	&	9752.0 	&	9526.1 	&	\textbf{90.0} 	\\	\hline
Pomeroy-2002-v1	&	9456.1 	&	9853.7 	&	9732.0 	&	10152.7 	&	9641.3 	&	9504.8 	&	\textbf{44.3} 	\\	\hline
Pomeroy-2002-v2	&	9695.7 	&	10643.3 	&	10351.3 	&	\textbf{9665.6} 	&	9910.3 	&	9862.0 	&	16729.5 	\\	\hline
Shipp-2002-v1	&	9709.8 	&	10286.8 	&	9979.0 	&	10539.6 	&	9975.2 	&	9901.0 	&	\textbf{163.0} 	\\	\hline
West-2001	&	9668.9 	&	10298.7 	&	10311.2 	&	10666.4 	&	10010.9 	&	9798.0 	&	\textbf{4049.5} 	\\	\hline
\textbf{Average}	&	9981.0 	&	10911.6 	&	10837.0 	&	10908.8 	&	10784.9 	&	10098.0 	&	\textbf{9099.5} 	\\	\hline
\end{tabular}}
\end{table*}

Table 8 shows the test accuracy of six EC methods and SaWDE on twelve higher-dimensional datasets. It can be seen that SaWDE can achieve the best results on 9 out of 12 datasets. Moreover, SaWDE can achieve the best performance in terms of average value, with an improvement of almost 13\% over the worst average result. Table 9 summarizes a feature subset size comparison of SaWDE with the six EC methods. We can observe that SaWDE is next to SaPSO among these methods, but its average subset size is smaller compared to the other methods.

Table 10 summarizes the results of those six EC methods and SaWDE in terms of CPU time. In our study, SaWDE ends the evolution early when 100\% accuracy is achieved in training and the feature subset size is less than half of the original subset. Therefore, from Table 10, we can observe that SaWDE can achieve the best performance in 9 datasets of 12. Moreover, the average CPU time of SaWDE is the best among all the compared algorithms.

\subsection{T-test Analysis of SaWDE and the other Methods}

This section provides T-test analysis of six EC method compared with SaWDE methods. Specifically, Tables 11 summarizes the T-test analysis significance level, where the results with less than 5\% indicate rejection of the null hypothesis at the 5\% significance level. We observe that SaWDE is significantly different from most algorithms in terms of subset size and training accuracy. For the test set, all EC methods showed significant differences from SaWDE. Therefore, we can conclude that our proposed SaWDE can provide significant results compared to other algorithms.

\begin{table*}[htb]
\centering
\caption{T-test analysis of six EC methods with SaWDE ($p$-value)}
\centering
\resizebox{120mm}{!}{
\begin{tabular}{|l |l| l| l| l| l| l|}
\hline
\textbf{T-rest } & \textbf{GA}  & \textbf{Original PSO} & \textbf{Standard PSO} & \textbf{SaPSO}  & \textbf{DE}  & \textbf{SaDE}\\
\hline 
\textbf{SaWDE (Subset Size)}	&	0.0012	&	0.0024	&	0.0022	&	0.6935	&	0.0076	&	0.0033	\\
\textbf{SaWDE (Training Accuracy)}	&	0.0607	&	0.0713	&	0.1238	&	0.6206	&	0.0084	&	0.0185	\\
\textbf{SaWDE (Test Accuracy)}	&	0.0024	&	0.0001	&	0.0002	&	0.0009	&	0.0016	&	0.0093	\\

\hline
\end{tabular}}
\end{table*}

\section{Conclusions}

In this article, we propose new approaches for DE to solve the feature selection problem, especially for large-scale feature selection. Notably, feature selection can be regarded as a combinatorial optimization problem, which requires a diversity of solutions to obtain a suitable result. Based on this fact, we use a multi-population to maintain the diversity of the population. Furthermore, a strategy pool is constructed to create various strategies and a self-adaptive strategy mechanism developed to  dynamically select the appropriate strategy based on the different characteristics of each dataset. Moreover, a weighted model is proposed to find the important features and then look for the best solutions by combining important features. Experiments were conducted on twelve datasets of varying difficulty. The experimental results show that SaWDE can achieve better performance in most cases in terms of classification accuracy than six non-EC methods and six other EC methods on both the training and test datasets. More importantly, the results show that the SaWDE algorithm could reduce 75\% to 80\% of the features in many cases and it shows high superiority compared to other EC methods on many datasets. What's more, the SaWDE algorithm shows excellent performance on large-scale datasets, which is encouraging for solving the problem of dimensional disaster.

Although we show that the SaWDE algorithm achieves promising results, more work is needed to improve the performance of the novel algorithm. For example, how to decide the number and size of the sub-populations, and indeed, whether equally divided sub-populations is a direction worth studying.

\bibliography{sample.bib}












\end{document}